\documentclass[final, 12p, twocolumn]{elsarticle}

\makeatletter
\def\ps@pprintTitle{%
  \let\@oddhead\@empty
  \let\@evenhead\@empty
  \let\@oddfoot\@empty
  \let\@evenfoot\@oddfoot
}
\makeatother

\usepackage{graphicx}
\usepackage{amssymb}

\usepackage{lineno}

\usepackage{floatrow}
\usepackage{hyperref}
\usepackage{subcaption}
\usepackage{multirow}

\journal{arXiv}

\begin{document}

\begin{frontmatter}


\title{Handgun detection using combined human pose and weapon appearance}



\author[label1]{Jesus Ruiz-Santaquiteria}
\author[label1]{Alberto Velasco-Mata}
\author[label1]{Noelia Vallez}
\author[label1]{Gloria Bueno}
\author[label2]{Juan A. Álvarez-García}
\author[label1]{Oscar Deniz}

\address[label1]{University of Castilla-La Mancha, ETSI Industriales, VISILAB, Ciudad Real, Spain}

\address[label2]{University of Seville, ETSI Informática, Dep. Lenguajes y Sistemas Informáticos, Sevilla, Spain}

\begin{abstract}

Closed-circuit television (CCTV) systems are essential nowadays to prevent security threats or dangerous situations, in which early detection is crucial. Novel deep learning-based methods have allowed to develop automatic weapon detectors with promising results. However, these approaches are mainly based on visual weapon appearance only. For handguns, body pose may be a useful cue, especially in cases where the gun is barely visible. In this work, a novel method is proposed to combine, in a single architecture, both weapon appearance and human pose information. First, pose keypoints are estimated to extract hand regions and generate binary pose images, which are the model inputs. Then, each input is processed in different subnetworks and combined to produce the handgun bounding box. Results obtained show that the combined model improves the handgun detection state of the art, achieving from 4.23 to 18.9 AP points more than the best previous approach.

\end{abstract}


\end{frontmatter}

\footnotetext[1]{This work has been submitted to the IEEE for possible publication. Copyright may be transferred without notice, after which this version may no longer be accessible}

\section{Introduction}
\label{S:1}

Video surveillance has come a long way in the past decades. Nowadays, public or private spaces such as train stations, airports, museums, banks or government institutional buildings have their own video surveillance systems. These systems are very useful for post-event investigations and also assisting the security personnel to manage crowds, being able to monitor different locations simultaneously. However, the main drawback of these solutions is the need for continuous monitoring by a human operator. The increasing number of areas controlled by video surveillance cameras, as well as factors inherent to human condition such as fatigue or loss of attention over time, make these systems rather inefficient~\cite{ainsworth2002buyer, velastin2006motion}.

Related studies in this area show that early detection of security threats or risks is fundamental to mitigate the damage caused as much as possible~\cite{enriquez2019vision}. Situations involving firearms such as handgun attacks, mass shootings, gunfire incidents on school grounds~\cite{gunfireUSA} or terrorist attacks~\cite{tessler2017use} are representative examples of this kind of threats, which unfortunately have become rather common nowadays.

The development of intelligent systems capable of automatically detecting threats or risk situations involving firearms as soon as possible can provide important advantages in terms of security. Recently, thanks to the momentum achieved by the introduction of deep learning methodologies, remarkable results have been obtained in visual tasks such as image classification or object detection and segmentation. In the particular case of firearm detection, while the results obtained with these novel methods are promising, there are still substantial limitations when they are applied in new scenarios different to those used for training, especially an unacceptable number of false positives~\cite{vallez2013false}. Furthermore, most of the proposed methods for automatic detection of firearms are based only on the appearance of the weapon in the image, without taking into account additional information that may help provide a more robust and accurate detection.

This work proposes the use of the human pose as complementary information to improve the performance of current handgun detectors based on deep learning. The human pose, defined as the relative position of the different joints and limbs of the human body, is quite characteristic in shootings. On the other hand, the images obtained by CCTV cameras are not generally of high quality, due to their low resolution, the presence of noise or poor lighting conditions. Also, other factors such as distance to the camera, the small visual size of the weapon\footnote{Powerful weapons like rifles and shotguns, which have a larger visual size, have not been considered in this work.} or a total or partial occlusion can prevent the object from being detected~\cite{castillo2019brightness}. Our hypothesis in this work is that contextual body information can help improve the robustness of the detection.

The contributions of this paper are as follows: (1) the development of a novel method for detecting hand-held firearms; (2) the performance evaluation of the proposed method in comparison with well-known appearance-based detection methods such as YOLOv3 and also other recent alternatives that consider human pose information and (3) assessment of the robustness of the method in environments under poor illumination conditions, with large camera distances and with different camera perspectives.

The rest of the article is organized as follows. Section~\ref{S:2} describes previous work related to the task of handgun detection. The datasets used in this study are detailed in Section~\ref{S:3}. In Section~\ref{S:4} the proposed method is explained. The experiments carried out and results obtained are summarized in Section~\ref{S:5}. Finally, conclusions and future work are presented in Section~\ref{S:6}.

\section{Related Work}
\label{S:2}

X-ray scanning machines are widespread in public spaces such as airports, train stations or museums with the objective of finding concealed weapons in luggage. The X-ray generated images are manually analyzed by a security operator. In this context, several approaches based on classical vision methods were proposed to automate the detection process. The work Nercessian
et al.~\cite{nercessian2008automatic} introduced a detection system based on image segmentation and edge-based feature vectors. Xiao et al.~\cite{xiao2015automatic} proposed a method based on Haar-like features and AdaBoost classifiers to automatically detect handguns in this kind of images. Also, 3D interest point descriptors have been studied for object classification in 3D baggage security computed tomography imagery~\cite{flitton2013comparison, flitton2015object}.

While X-ray imaging-based systems are useful to find weapons in travel bags or luggage, the scope of these solutions is very limited. Additionally, this kind of scanning machines are quite expensive. Using the RGB images captured by the CCTV video surveillance cameras to detect dangerous objects can be a more versatile and economical option. In this respect, several works related to the detection of weapons in RGB images through traditional machine learning methods have been proposed. Tiwari and Verma~\cite{tiwari2015computer} proposed a method to detect weapons in RGB images which used color segmentation and the k-means algorithm to remove unrelated objects. Then, the Harris interest point detector and Fast Retina Keypoint (FREAK) is used to locate the handgun in the segmented images. Later, Halima and
Hosam~\cite{halima2016bag} proposed another method to detect the presence of a handgun in an image. In this case, SIFT features are extracted from the collection of images and clustered by the k-means algorithm. Then, a word vocabulary based histogram is implemented and finally, a Support Vector Machine is used to decide whether the new image contains a weapon.

More recent deep learning based methods have been also applied to this task using different strategies. An important family of them is based on sliding windows. In this case, a large number of regions or windows of different sizes and aspect ratios are generated within the image (on the order of $10^{4}$) and each one is classified individually by a neural network. Several studies have applied this technique to detect handguns in images similar to those captured by CCTV video surveillance cameras~\cite{grega2016automated, gelana2019firearm}. The major drawback of this type of system is the high processing time required to classify these windows,  making it difficult to use them in real time. Other solutions are based on region proposals, which instead of using all possible windows in an image select only actual candidates. The first techniques that have used CNNs in this context are the Region-based CNN family of methods~\cite{girshick2015fast, ren2015faster}. Verma and Dhillon~\cite{verma2017handheld} proposed a method based on the Faster-RCNN framework with a VGG-16 backbone as feature extractor trained with the IMFDB dataset~\cite{imfdb} to detect hand-held arms. Both sliding window and Faster-RCNN methods were tested and compared by Olmos et al.~\cite{olmos2018automatic} for handgun detection. Faster-RCNN pre-trained with VGG-16 architecture obtained the best results on a custom dataset of 3000 YouTube gun images. Finally, another common approach for detecting objects is based on the YOLO family of methods~\cite{redmon2016you, redmon2017yolo9000,farhadi2018yolov3}. In these architectures a single deep neural network is applied once to the full image instead of multiple region proposals. The image is divided into fixed regions and probabilities and bounding boxes are predicted for each one. Several works have also recently applied YOLOv3 for detecting firearms with promising results~\cite{warsi2019gun,de2019firearm}.

Human pose information has been recently used for handgun detection and threat assessment. Abruzzo et al.~\cite{abruzzo2019cascaded} proposed a method for identifying people and handguns in images and then evaluate the threat level of the person poses based on the their body posture. However, the main limitation of this work is that the handgun detection performance is limited by the handgun detector used (in this case YOLO). In the handgun detection step, no human pose information is considered. Basit et al.~\cite{basit2020localizing} proposed a method for classifying human-handgun pairs. As in the previous work, human and handgun are separately detected. Then, each detected human is paired with each detected handgun and, finally, a neural network is trained to classify these paired human-handgun bounding boxes into two classes: ``carrying handgun'' and ``not carrying handgun''. This method can be used to remove false handgun detections, but again the detection performance is limited by the handgun detector used and the human pose cannot help to reduce the number of false negatives. More recently, Salido et al.~\cite{salido2021} analyzed how including body pose information (skeleton keypoints and limbs retrieved by a pose detector) in the input images, as a preprocessing step, can improve the handgun detection performance.

In the closest work to ours, an approach to improve a handgun detector through the integration with the human pose was introduced recently in Velasco-Mata et al.~\cite{velasco2021human}. This method used a visual heatmap representation of both the pose and the weapon location, using convolutional layers to obtain a final grayscale image that indicates potential handgun regions on the image.

\section{Materials}
\label{S:3}

This section describes the datasets used for assessing the performance of the proposed method. In order to consider different contexts and image features, the images have been collected from different sources, such as public handgun datasets, YouTube clips and even synthetic images obtained from video games. 

\subsection{Public handgun datasets}
\label{S:3.1}

\begin{figure}[htbp]
    \centering
    \begin{subfigure}{0.48\textwidth}
        \includegraphics[width=\textwidth]{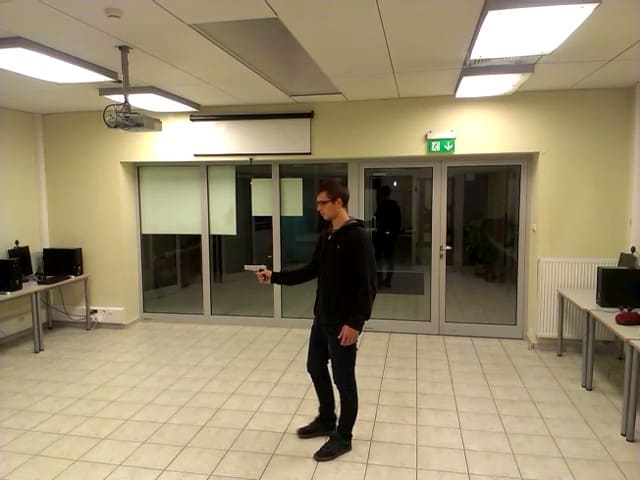}
    \end{subfigure}
    \begin{subfigure}{0.48\textwidth}
        \includegraphics[width=\textwidth]{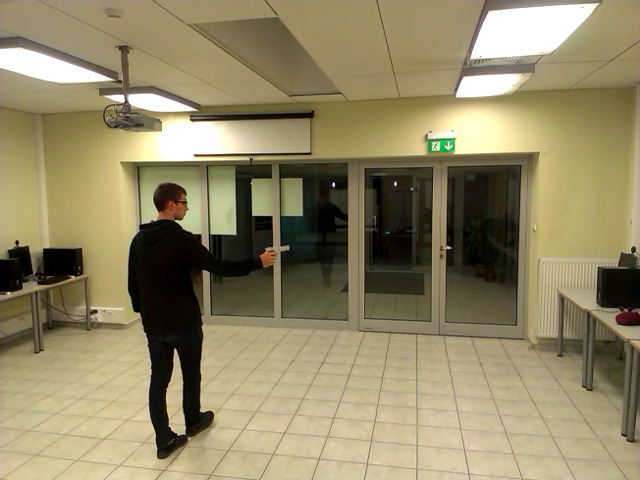}
    \end{subfigure}
    \caption{Sample images from Gun Movies Database}
    \label{fig:dataset_GMD_samples}
\end{figure}

The proposed method is intended to be applied in CCTV surveillance systems on a wide variety of scenarios. Unfortunately, most public handgun datasets contain weapon profile images occupying the whole image and with homogeneous background, which are quite different from the type of images captured with surveillance cameras. Surveillance scenarios are typically characterized by a large distance between the subjects recorded and the camera, low image resolution or quality, and poor lighting conditions, among others. Salazar Gonzalez et al.~\cite{SalazarGonzalez2020} recently introduced a new dataset composed of CCTV images from a real video surveillance system and synthetic images generated with the Unity game engine. However, the CCTV images in this dataset are not realistic enough, showing unnatural poses for handgun attacks or mass shootings. On the other hand, it is possible to find some public datasets or parts thereof which are realistic enough. The first dataset used in our study is composed of 665 images of size 640x480 extracted from videos of the Guns Movies Database~\cite{grega2016automated}. In these clips a man is holding a handgun in a few shooting poses in an indoor room. Camera distance, image resolution and illumination conditions are a good representation of CCTV scenarios. In~\autoref{fig:dataset_GMD_samples} two example images from this dataset are shown.

\begin{figure}[htbp]
    \centering
    \begin{subfigure}{0.48\textwidth}
        \includegraphics[width=\textwidth]{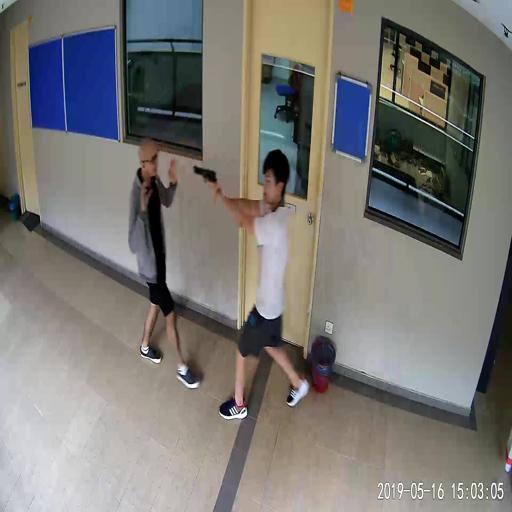}
    \end{subfigure}
    \begin{subfigure}{0.48\textwidth}
        \includegraphics[width=\textwidth]{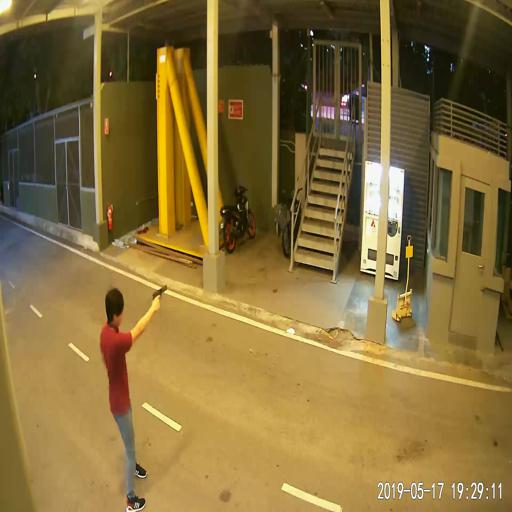}
    \end{subfigure}
    \caption{Sample images from Monash Guns Dataset}
    \label{fig:dataset_MGD_samples}
\end{figure}

Also, 300 images of size 512x512 were obtained from the publicly available Monash Guns Dataset~\cite{lim97deep} for test purposes. These images show different CCTV scenarios with people holding handguns in several body poses. In~\autoref{fig:dataset_MGD_samples} two example images from this realistic dataset are shown.

\subsection{YouTube videos}
\label{S:3.2}

\begin{figure}[htbp]
    \centering
    \begin{subfigure}{0.48\textwidth}
        \includegraphics[width=\textwidth]{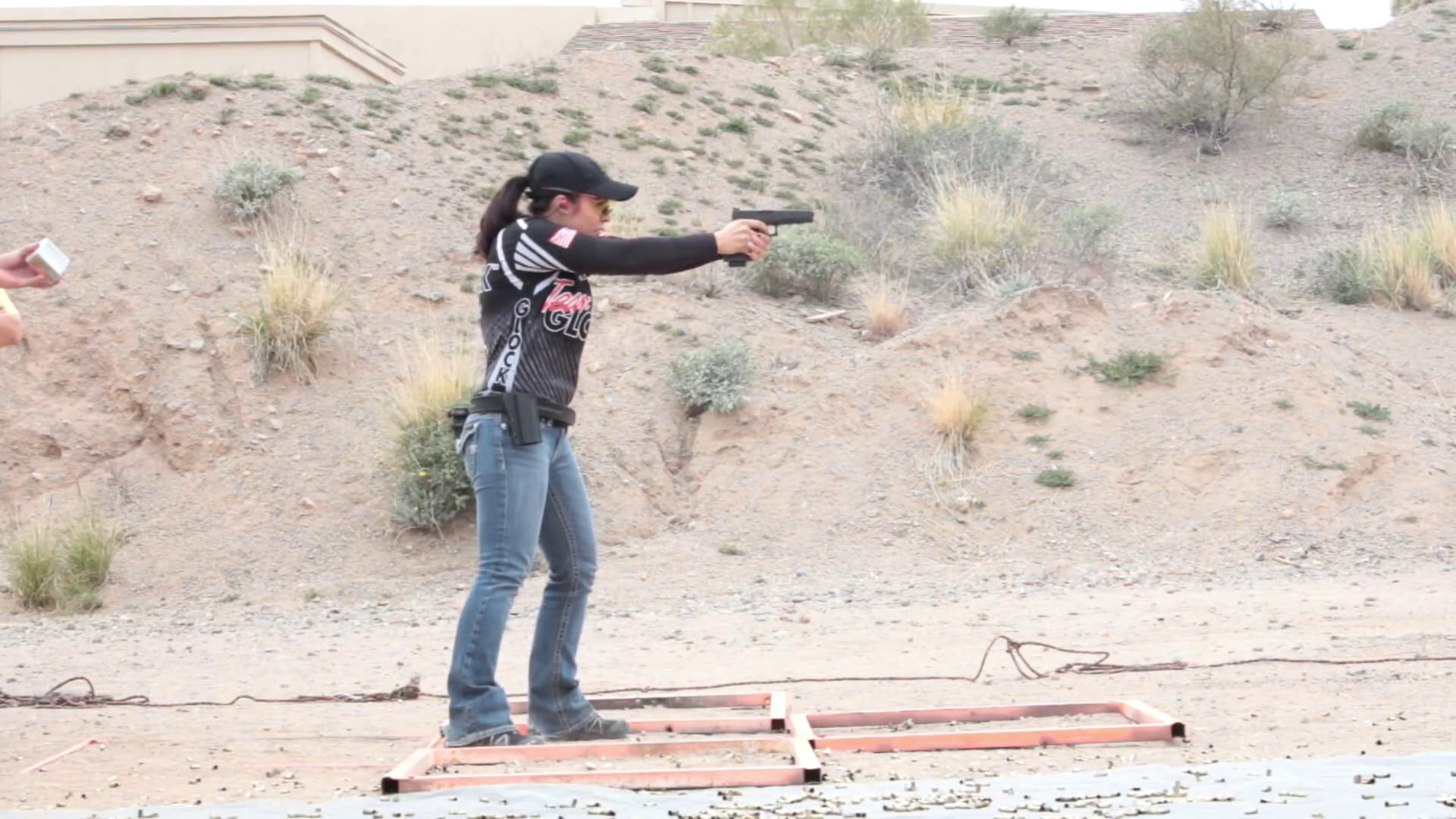}
    \end{subfigure}
    \begin{subfigure}{0.48\textwidth}
        \includegraphics[width=\textwidth]{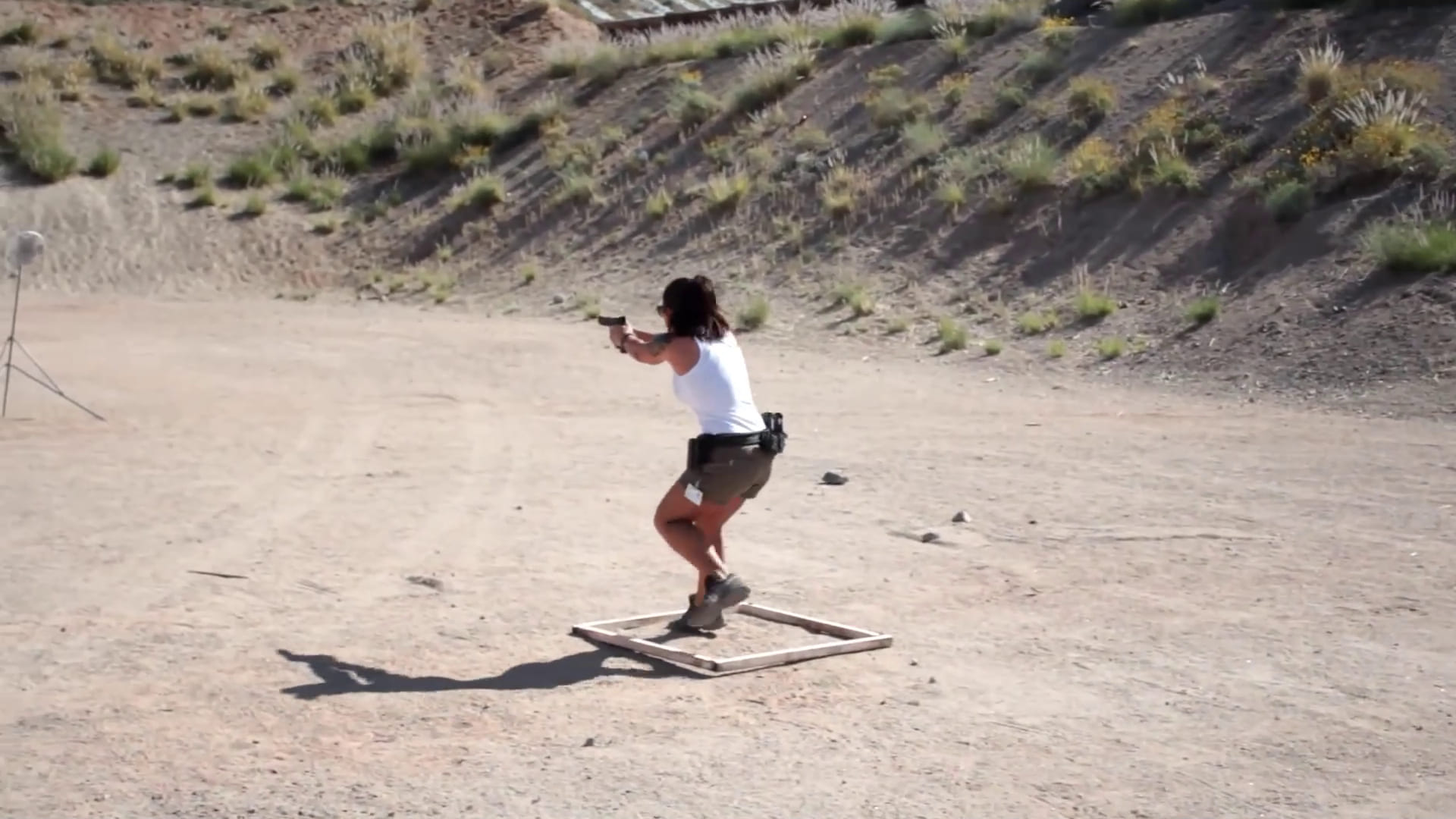}
    \end{subfigure}
    \caption{Sample images from YouTube dataset}
    \label{fig:dataset_YouTube_samples}
\end{figure}

YouTube is another useful source to find videos of people carrying or holding weapons and/or shooting. As in the previous case, it is difficult to find clips of real CCTV footage showing handguns. Nevertheless, there are videos of shooting practice sessions which are suitable for our purposes. This dataset is composed of 952 images of size 1920x1080 extracted from 12 YouTube clips. In these videos there are different camera locations, background scenarios, shooting poses and lighting conditions.~\autoref{fig:dataset_YouTube_samples} shows two example images from this dataset.

\subsection{Synthetic video game images}
\label{S:3.3}

\begin{figure}[htbp]
    \centering
    \begin{subfigure}{0.48\textwidth}
        \includegraphics[width=\textwidth]{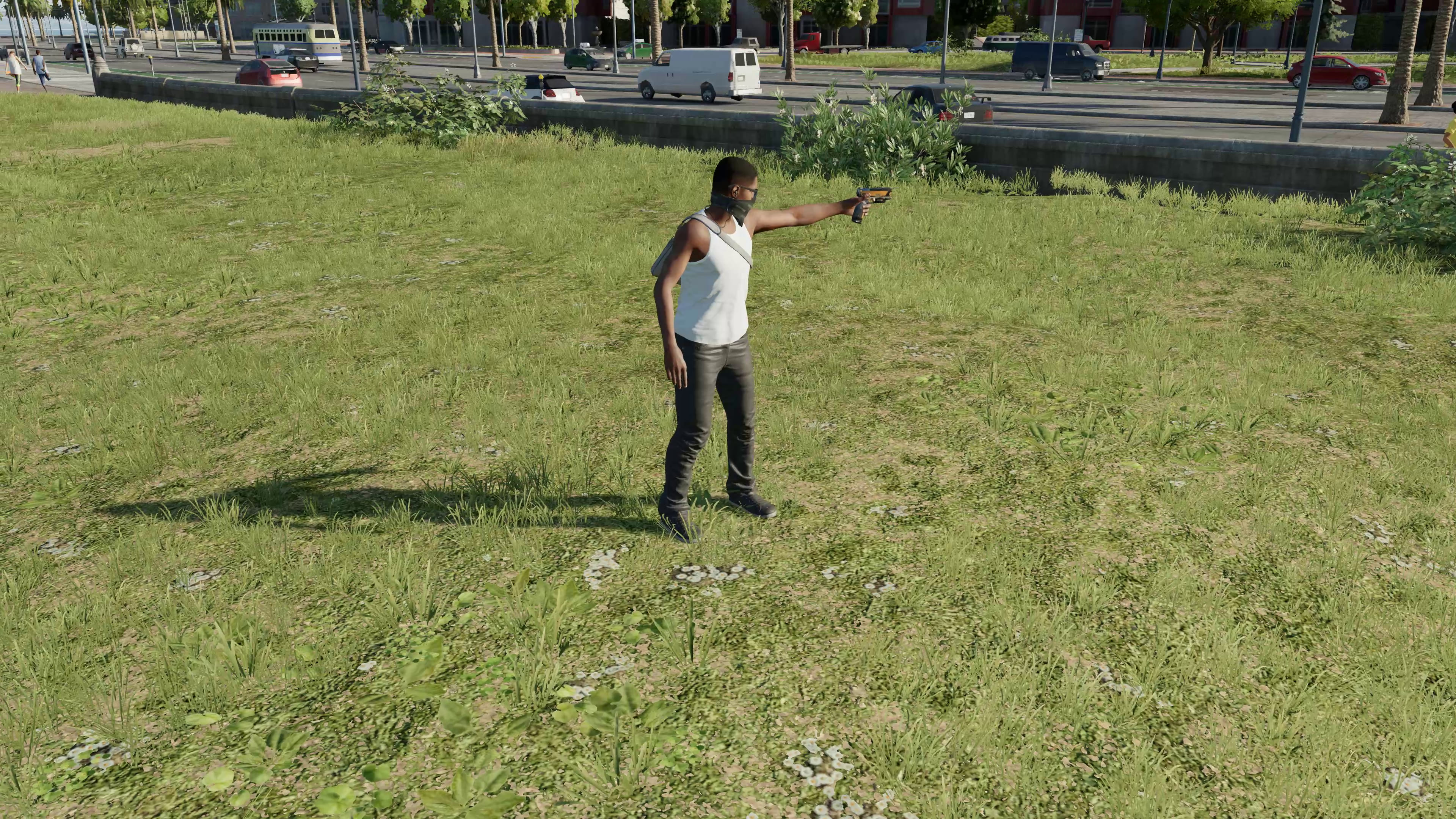}
    \end{subfigure}
    \begin{subfigure}{0.48\textwidth}
        \includegraphics[width=\textwidth]{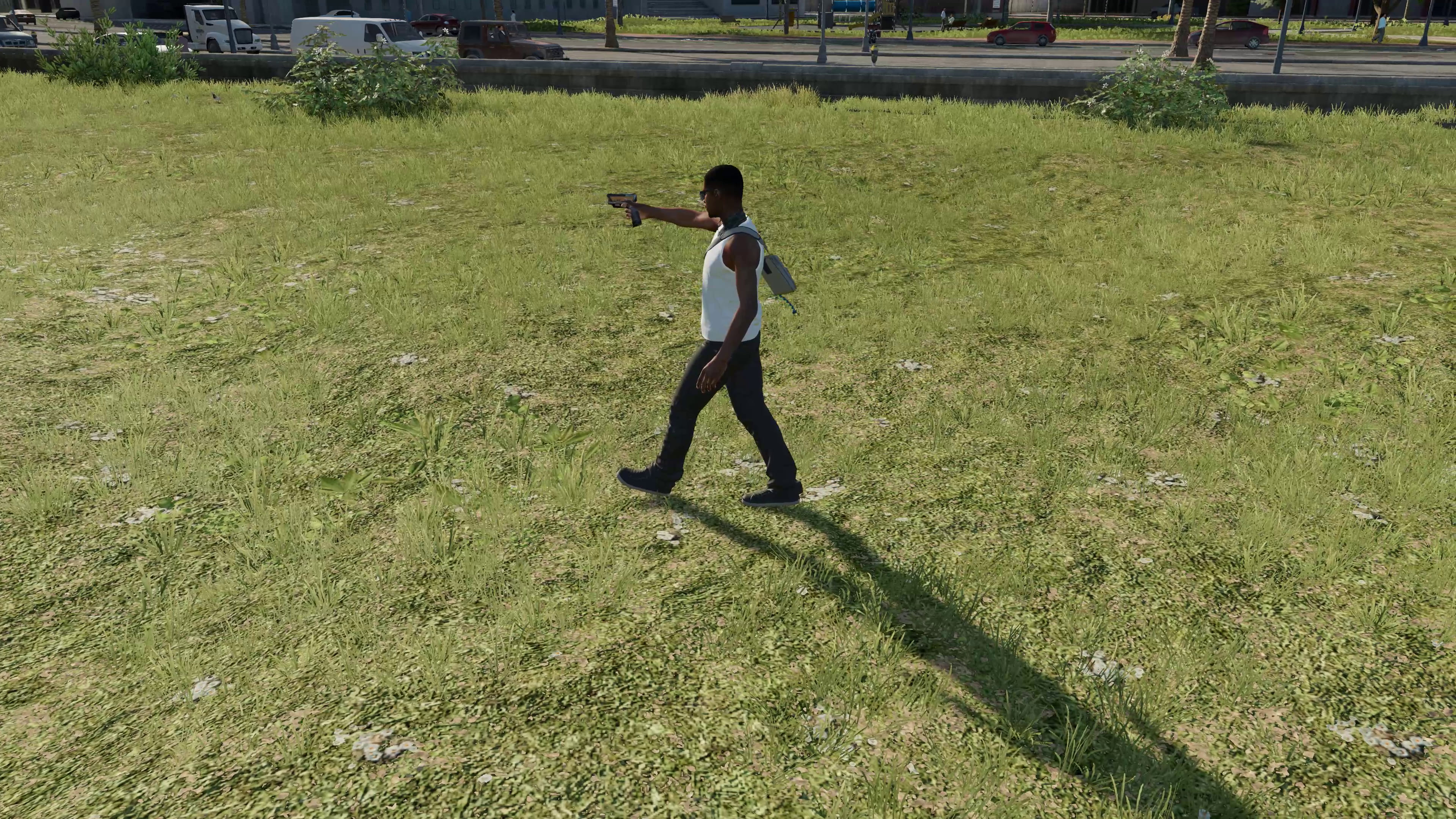}
    \end{subfigure}

    \caption{Sample images from Watch Dogs 2 dataset (© 2016 Ubisoft Entertainment)}
    \label{fig:dataset_WD2_samples}
\end{figure}

Video games can be also used to create new data for this task. Through specific video games it is possible to recreate representative situations or scenarios and then extract videos or images. In this case, a synthetic dataset was created with the popular shooter video game Watch Dogs 2 on a PC platform. Using the novel NVIDIA Ansel feature\footnote{\url{https://developer.nvidia.com/ansel}}, ingame videos can be recorded from different camera locations, distances or angles. In this way, 4 video sequences were recorded, performing a full camera rotation around the main character with two different heights in various shooting animations. Finally, 650 images of size 3840x2160 were obtained from these video sequences. In~\autoref{fig:dataset_WD2_samples} two example images of this dataset are shown.

\subsection{Data augmentation and dataset split}
\label{S:3.4}	

The use of a large and representative dataset is essential to achieve good performance in novel object detectors based on deep learning and CNNs. YouTube clips or synthetic video game images can be a solution, but the manual labelling process is a time consuming task, limiting the number of images that can be effectively generated.

Novel deep learning methods need a huge amount of data to be correctly trained, because of the large number of parameters and model complexities. To deal with this problem, data augmentation is a common practice that helps to increase the size and variability of the dataset by applying a set of transformations to the original data. In our case, for each image included in the dataset, another one was generated performing a horizontal flip. In this way, the number of images was doubled, generating new shooting poses. 

Moreover, transfer learning is another technique commonly used in classification or object detection tasks to circumvent the need for large training datasets. It is based on adapting the useful features learned in a previous training process with a large public dataset, such as COCO~\cite{coco} or ImageNet~\cite{imageNet}, to a new domain. In this way, starting the new training with the previously learned parameters instead of randomly initialized weights, allows us to obtain good results for the specific task even with small specific datasets.	

Finally, for this work, the dataset used for training the proposed method after the data augmentation procedure is composed of a total of 3000 randomly selected images, obtained from Guns Movies Database, Watch Dogs 2 and YouTube datasets, including a total of 3160 handguns. The validation set is formed by a total of 300 images obtained from the same datasets (Guns Movies Database, Watch Dogs 2 and YouTube), including 306 handguns. For test, two different sets have been created. The first one (test set A) is formed by a total of 300 images of the YouTube database, including a total of 297 handguns. The second one (test set B) is composed of 300 images from the Monash Guns Dataset, containing another 300 handguns. None of the test images was either in the training or validation set. A summary of the dataset composition is presented in~\autoref{tab:tableDataset}.

\begin{table*}[ht]
\centering
\caption{Number of images for each dataset after the data agumentation process.}
\label{tab:tableDataset}
\begin{tabular}{|l|c|c|c|c|c|}
\cline{1-6}
                       & \textbf{Training} & \textbf{Validation}   & \textbf{Test set A}   & \textbf{Test set B}   & \textbf{Total}   \\ \cline{1-6}
Gun Movies Database    & 1064              & 113                   &   0                   &   0                   & \textbf{1177}              \\ \cline{1-6}
Monash Guns Dataset    &    0              &   0                   &   0                   & 300                   &  \textbf{300}              \\ \cline{1-6}
Watch Dogs 2           & 1071              & 108                   &   0                   &   0                   & \textbf{1179}              \\ \cline{1-6}
YouTube                &  865              &  79                   & 300                   &   0                   & \textbf{1244}              \\ \cline{1-6}
\textbf{Total}         & \textbf{3000}     & \textbf{300}          & \textbf{300}          & \textbf{300}          & \textbf{3900}      \\\cline{1-6}  
\end{tabular}
\end{table*}

\section{Methodology}
\label{S:4}

In this section, the different steps involved in the proposed method are detailed, starting from the input image down to the final handgun detections.

\subsection{Human pose estimation}
\label{S:4.1}

The first step consists of collecting the human pose information found in the input image. This was done with the OpenPose framework~\cite{cao2019openpose}. OpenPose is an open-source multi-person pose estimator which is able to predict the 2D keypoints as well as keypoint associations, keeping a high accuracy and a low inference time. In this step, a set of 25 2D keypoints are predicted for each person in the image, along with predicted confidence for each one of them. These keypoints include the necessary human body position information (neck, shoulders, elbows, wrists, etc.) to define the pose of each person.

\subsection{Hand region extraction}
\label{S:4.2}

\begin{figure*}[htbp]
	\centering
	\includegraphics[width=0.95\textwidth]{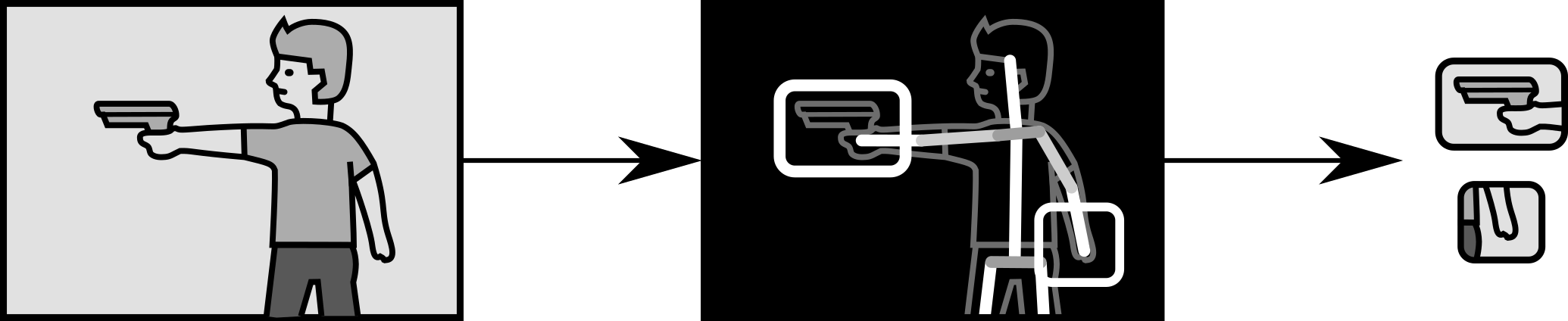}
	\caption{Pose estimation and hand region extraction}
	\label{fig:poseHandEstimation}
\end{figure*}

In the second step, using the collected pose information, the hand regions for each detected person are inferred and extracted. The elbow and wrist positions, as well as the distances and directions between them are used to generate a set of bounding boxes around all of the hand regions in the input image (see~\autoref{fig:poseHandEstimation}).

The confidence score given by OpenPose for each keypoint is applied to filter wrong or inaccurate detections, and an intersection over union (IoU) threshold between the predicted bounding boxes is checked to prevent overlapping areas (e.g., a handgun held with both hands is considered as a single region, since both bounding boxes are overlapping). 

\subsection{Hand region classification}
\label{S:4.3}

\begin{figure*}[htbp]
	\centering
	\includegraphics[width=0.8\textwidth]{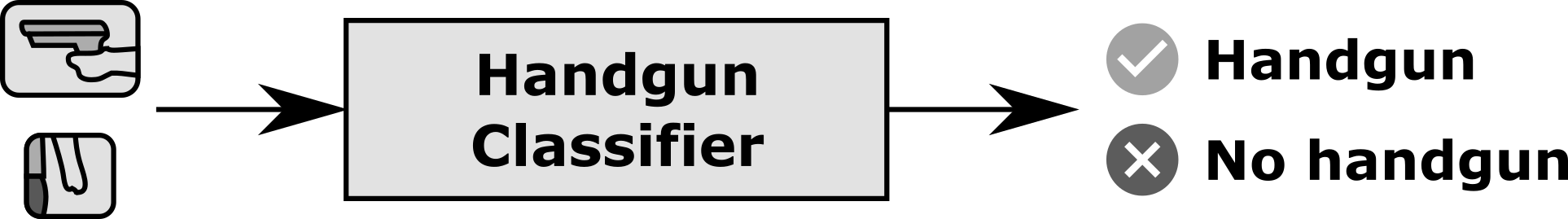}
	\caption{Hand region classification step}
	\label{fig:handClassification}
\end{figure*}

For this stage, a convolutional neural network was trained to classify the previously generated hand regions into handgun or no-handgun areas, based on whether there is a handgun inside the region (see~\autoref{fig:handClassification}). The selected network was Darknet-53, the backbone feature extractor used in the YOLOv3 object detector. Henceforth, this hand region classifier will be denoted as HRC (Hand Region Classifier).

The dataset used for training the hand region classifier was composed of 6177 images, generated from the hand areas extracted from the 3000 training images described in Section~\ref{S:3}. These regions were automatically labelled by comparing the hand areas and the ground truth handgun locations. For this, instead of the IoU score, we followed the overlap measure proposed in Velasco-Mata et al.~\cite{velasco2021human}: intersection over minimum area (IoMin), see~\autoref{eq:iomin}. Usually, ground truth handgun locations are smaller than hand bounding boxes. This metric allows a better overlap measurement in this particular scenario, since bounding boxes of different sizes are not penalized. More details are given in~\cite{velasco2021human}.

\begin{equation}
	IoMin(A,B) = \frac{area(A \cap B)}{min(area(A), area(B))}
\label{eq:iomin}
\end{equation}

If the regions overlap with a 0.5 IoMin threshold, the hand area is labelled as a handgun area. On the other hand, if there is no overlap or the treshold is below 0.5 IoMin, the hand area is labelled as no-handgun. 

Each hand region was also resized to a fixed size of 256x256. The model was trained with a batch size of 4 in 60 epochs using the Adam optimization algorithm and the categorical cross entropy as loss function.

\subsection{Pose combination method}
\label{S:4.4}

A further modification of the HRC method described in the previous section was considered. In this case, the network is modified to combine the hand region image with the human pose information obtained with OpenPose. This was done to help the region classifier by exploiting correlation with the individual's pose information. Pose data is used to create binary images of fixed size 512x512 for each detected person in the input image, drawing the keypoints and the connections between them. A normalization procedure is also applied to focus only on the relative position between the keypoints, removing variable factors such as camera distance and absolute position in the image. For this, the original neck keypoint $j_0$ is taken as reference and the distance between this point and the lumbar spine keypoint $j_1$ is used as the scale factor for the normalization. In this way, the new keypoints $k_n$ are calculated following~\autoref{eq:localCoordinates}:

\begin{equation}
	k_n = \frac{j_n - j_0}{|\overrightarrow{j_0j_1}|}
	\label{eq:localCoordinates}
\end{equation}

where $j_n$ is the original 2D point and $|\overrightarrow{j_0j_1}|$ is the distance between $j_0$ and $j_1$.

\begin{figure*}[htbp]
	\centering
	\begin{subfigure}[c]{0.48\textwidth}
		\includegraphics[width=\textwidth]{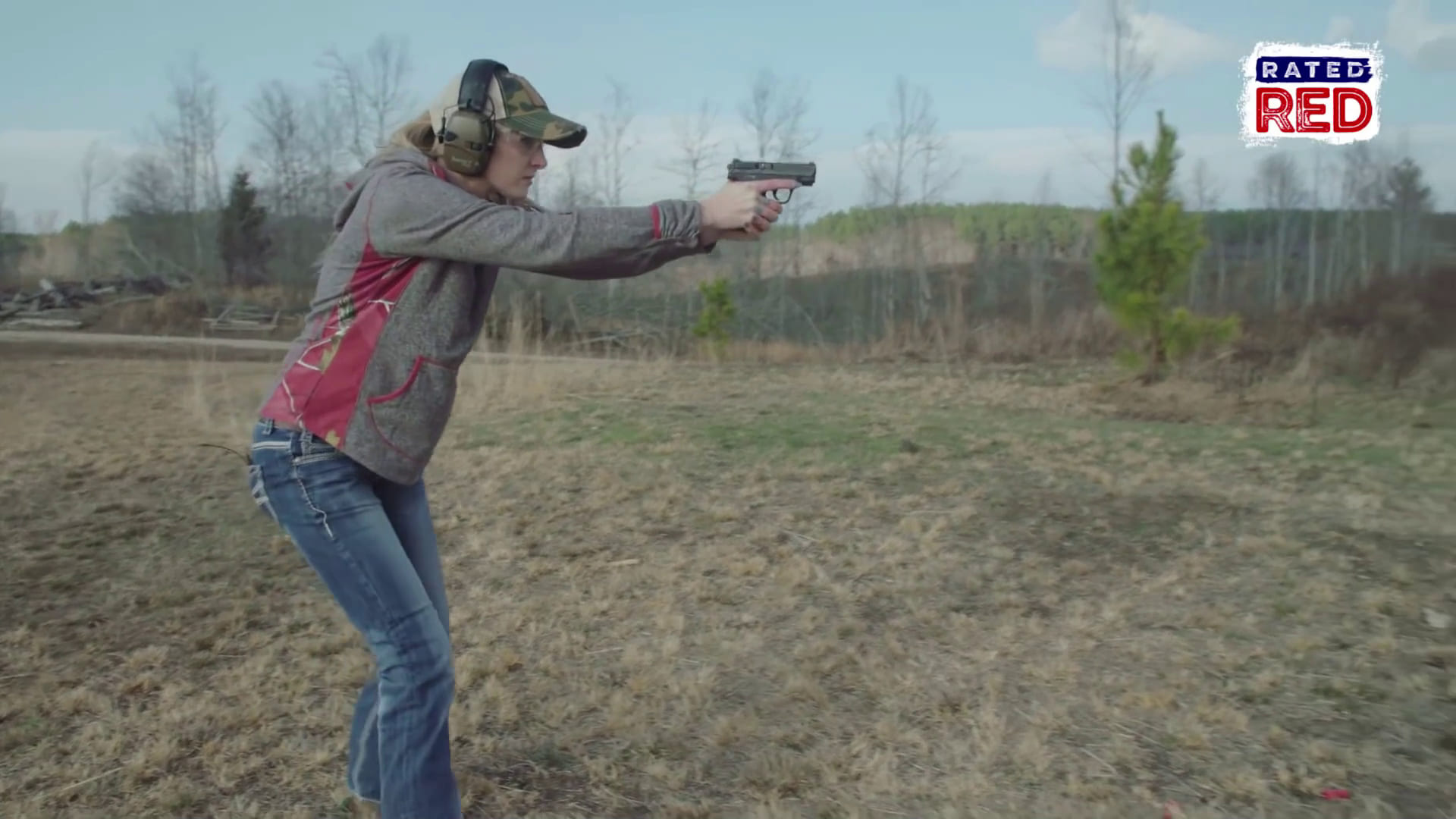}
		\caption{Original OpenPose image}
		\label{fig:poseClassificationExample_a}
	\end{subfigure}\hspace{12pt}
	\vspace{1pt}
	\begin{subfigure}[c]{0.27\textwidth}
		\includegraphics[width=\textwidth]{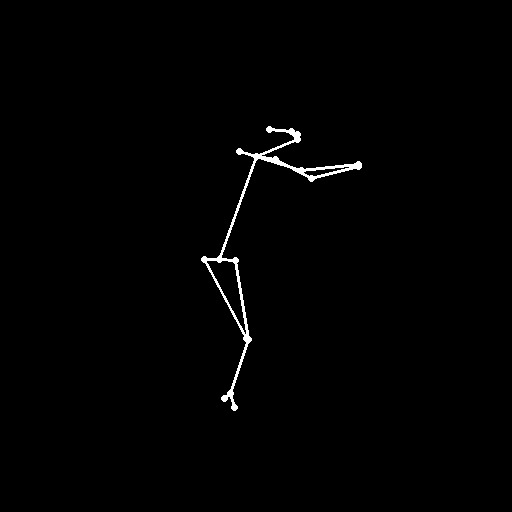}
		\caption{Normalized pose image}
		\label{fig:poseClassificationExample_b}
	\end{subfigure}
	\caption{Pose classification example.}
	\label{fig:poseClassificationExample}
\end{figure*}

\begin{figure*}[htbp]
	\centering
	\includegraphics[width=0.9\textwidth]{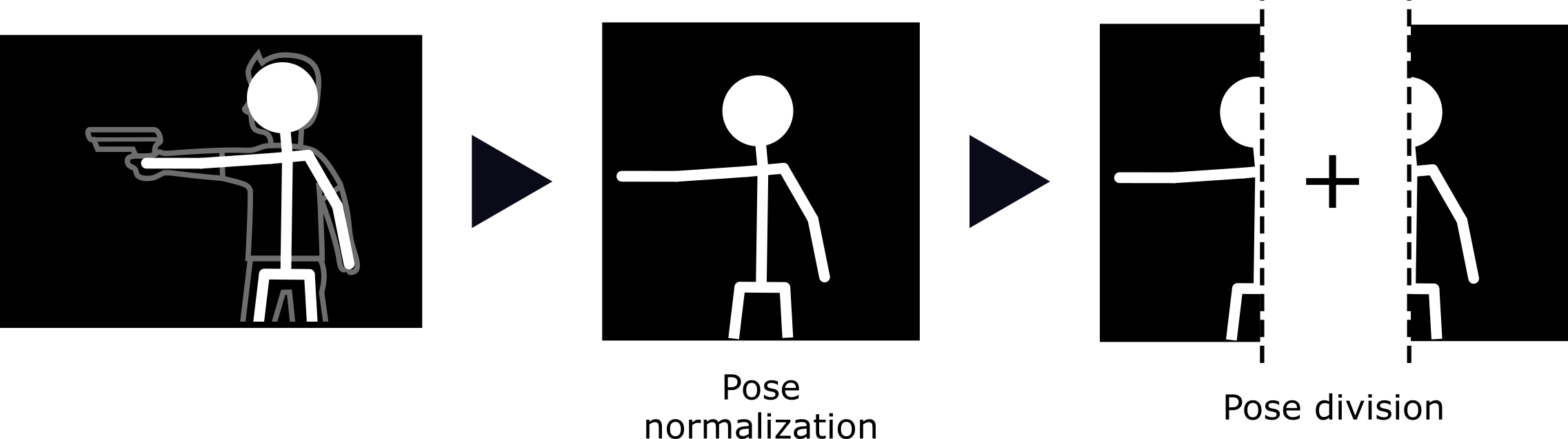}
	\caption{Pose processing steps, including normalization and division in halves}
	\label{fig:poseProcessing}
\end{figure*}

In~\autoref{fig:poseClassificationExample_b} an example of a generated pose image is shown, along with the original image (\autoref{fig:poseClassificationExample_a}). This binary pose image along with the original hand region image are the inputs to the new classifier. However, note that each pose image is related to two different hand regions (the two hands of an individual), and the pose image generated would be the same in the two cases. The problem then is that one of these regions could be labelled as handgun and the other as no-handgun, and the net effect of the additional pose information would be ignored (as the two regions have the same pose image). To prevent this, the pose image is divided into two parts of size 256x512 as shown in Figure~\ref{fig:poseProcessing}, selecting as input the half in which the hand region is included. Thus, for each detected hand region there are two network inputs, the hand region itself and the pose image half corresponding to this hand region. For those cases in which the handgun is held with the two hands and the bounding boxes are overlapping, as the case shown in Figure~\ref{fig:poseClassificationExample}, we ensure that only a single hand region is taken into account, along with the corresponding pose image half. 

\begin{figure*}[htbp]
	\centering
	\includegraphics[width=0.90\textwidth]{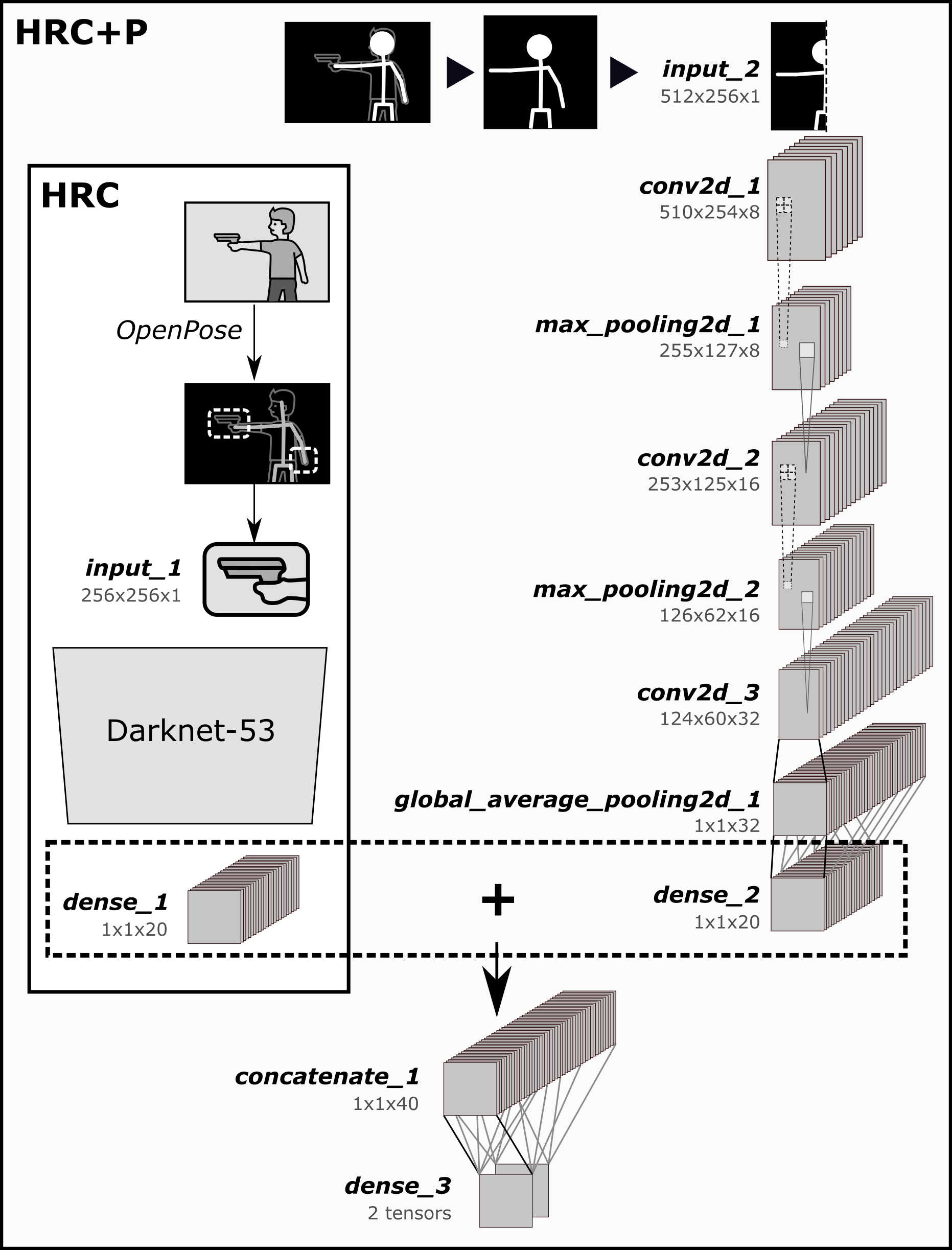}
	\caption{Network architecture}
	\label{fig:networkArchitecture}
\end{figure*}

The whole network architecture, henceforth named as HRC+P (Hand Region Classifier + Pose data), is divided into two main branches. The first one is the hand region classifier (HRC). On the other hand, the processing of the pose image is carried out by another custom subnetwork. The outputs of the two subnetworks are then joined in a single feature vector connected to the output of the whole network. In this way, the model is capable of learning the optimal combination of handgun appearance with human pose information to improve the classification performance. The whole architecture is shown in~\autoref{fig:networkArchitecture}. 

\subsection{Bounding box prediction}
\label{S:4.5}

\begin{figure*}[htbp]
	\centering
	\includegraphics[width=0.7\textwidth]{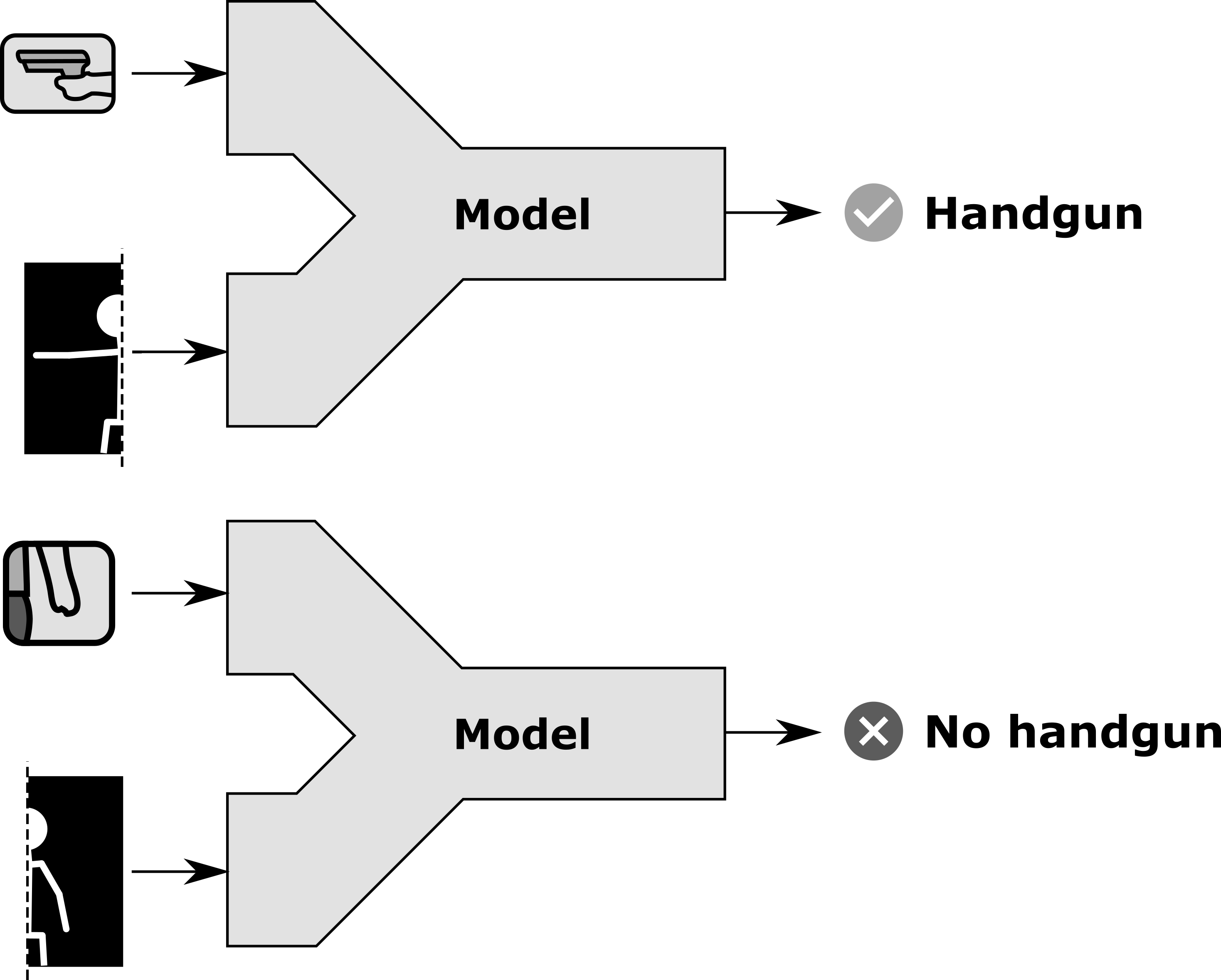}
	\caption{Bounding box prediction step}
	\label{fig:bboxPrediction}
\end{figure*}

The last step of the proposed method consists of generating the handgun predictions in the image. Each hand region of each detected person is passed through the classification network to obtain a class label (handgun \textit{vs} no-handgun). Then, the bounding boxes of the regions classified as handgun are included in the output list of predicted handguns (see~\autoref{fig:bboxPrediction}). 

\section{Results}
\label{S:5}

This section presents the results obtained in the tests carried out to evaluate the performance of the proposed method. In general, object detection models are evaluated using Precision, Recall, and Average Precision (PASCAL VOC AP50) metrics~\cite{padillaCITE2020}. In fact, these metrics are based on True Positives (TP), False Positives (FP) and False Negatives (FN). These values are calculated taking into account the overlap between the ground truth bounding boxes and those predicted by the detector. In the same way as in the automatic labeling process for the training of the hand region classifier (Subsection~\ref{S:4.3}), the IoMin is the selected criterion to calculate the overlap between the predicted bounding boxes and the ground truth data, due to the size difference between them.

The proposed pose-combined approach (HRC+P) has been compared to three different handgun detectors:
\begin{itemize}
	\item YOLOv3~\cite{farhadi2018yolov3}: YOLOv3 is one of the fastest and most accurate deep learning-based object detectors. The Darknet-53 CNN backbone is used as feature extractor, which provides an interesting baseline for comparison. 
	
	\item Basit et al.~\cite{basit2020localizing}: This work proposes a method to classify person-handgun pairs detected in an image, between people carrying handguns and those who do not. 
	
	\item Velasco-Mata et al.~\cite{velasco2021human}: This recent work, the most similar to ours, proposes a method to improve a handgun detector based on a visual heatmap representation of both pose and weapon location. 
	
	\item Salido et al.~\cite{salido2021}: This method overlays body pose information retrieved by the OpenPose framework to the input images. In this way, CNN-based detectors can learn the association of a handgun location with the visual patterns of the pose skeletons included in the images.
	
	\item The proposed hand region classifier without pose information (HRC): To check the effect of including the 2D human pose information in the hand region classifier, the hand region processing branch without pose combination is taken for comparison.
\end{itemize}

All methods were trained and tested using the datasets described in Section~\ref{S:3}. 

\subsection{Test set A - Original data}

\begin{figure}[!htb]
	\centering
	\includegraphics[width=0.9\textwidth]{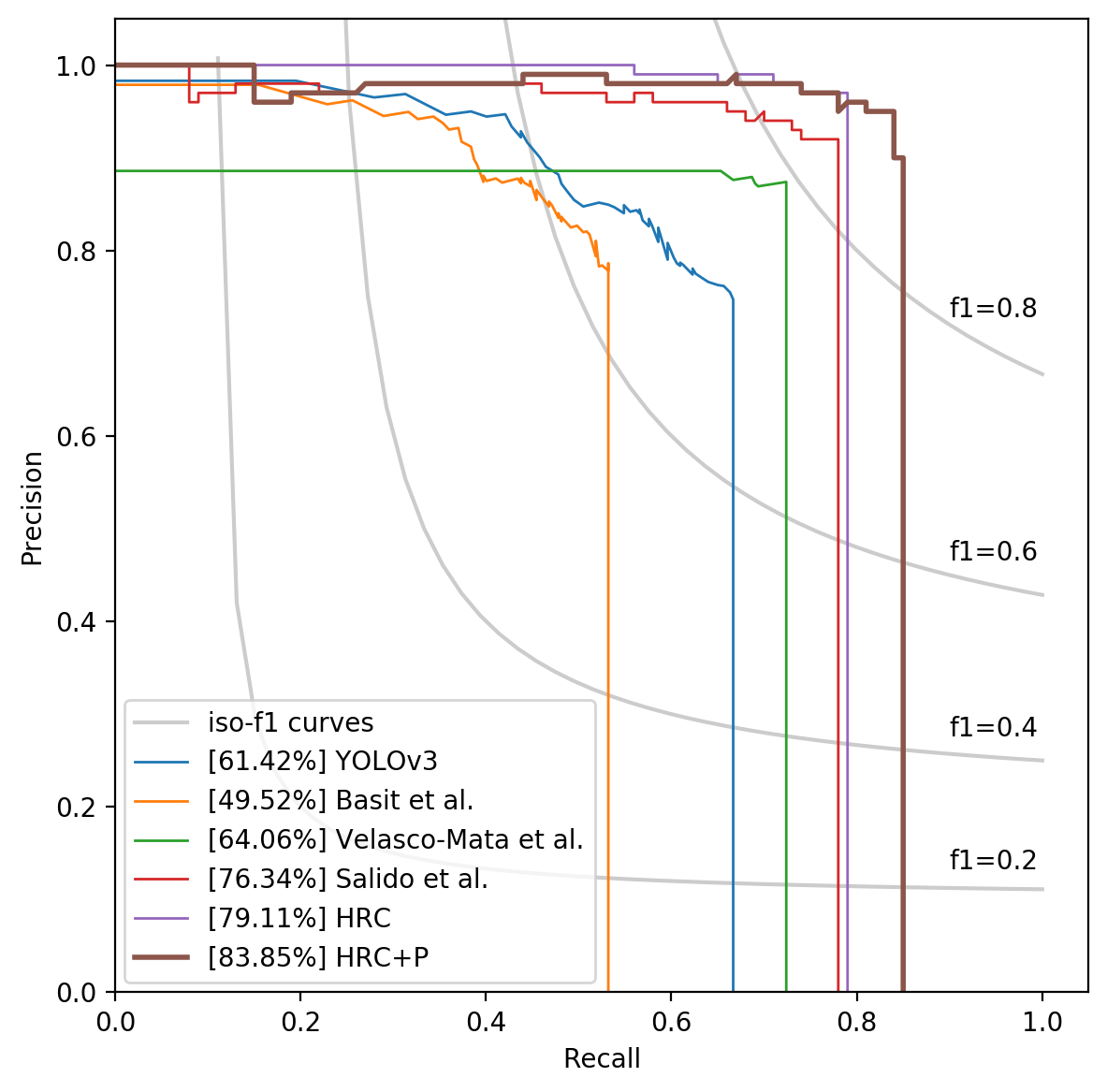}
	\caption{Precision-Recall curves obtained for test set A with the original images.}
	\label{fig:apOriginal}
\end{figure}

The results obtained in the test set A are summarized in~\autoref{tab:tableResults_original}. Precision and Recall values are calculated with a 0.5 prediction score threshold. Also, Precision-Recall curves of all methods are shown in~\autoref{fig:apOriginal}.


\begin{table*}[ht]
\caption{Evaluation metrics for test set A with the original images.}
\label{tab:tableResults_original}
\vskip 0.1in
	\begin{center}
		\begin{tabular}{|c|c|c|c|}
			\cline{1-4}
			\textbf{Method}                                     & \textbf{Precision 0.5} & \textbf{Recall 0.5} & \textbf{AP}    \\  \cline{1-4}
			\textbf{YOLOv3~\cite{farhadi2018yolov3}}            & 0.7937                 & 0.5959              & 61.42          \\
			\textbf{Basit el al.~\cite{basit2020localizing}}    & 0.8544                 & 0.4545              & 49.52          \\
			\textbf{Velasco-Mata et al.~\cite{velasco2021human}}       & 0.8692                 & 0.6936              & 64.06          \\
			\textbf{Salido el al.~\cite{salido2021}}            & 0.9351                 & 0.7272              & 76.34          \\
			\textbf{HRC}                                        & \textbf{0.9733}        & 0.7374              & 79.11          \\
			\textbf{HRC+P}                                      & 0.9188                 & \textbf{0.8383}     & \textbf{83.85} \\
			\cline{1-4}
		\end{tabular}
	\end{center}
\end{table*}

The highest AP score is achieved with the HRC+P method, the pose-combined version of the proposed approach. The AP obtained is approximately 5\% higher than the no-combined version (HRC). HRC+P is able to detect the largest number of handguns in the dataset. However, the least number of false positives is obtained with the HRC approach. 

To better assess the performance of the models under different conditions, two additional versions of the test set A were generated. The results are described in what follows.

\subsection{Test set A - Dark data}

\begin{figure*}[ht]
	\centering
	\begin{subfigure}[c]{0.48\textwidth}
		\includegraphics[width=\textwidth]{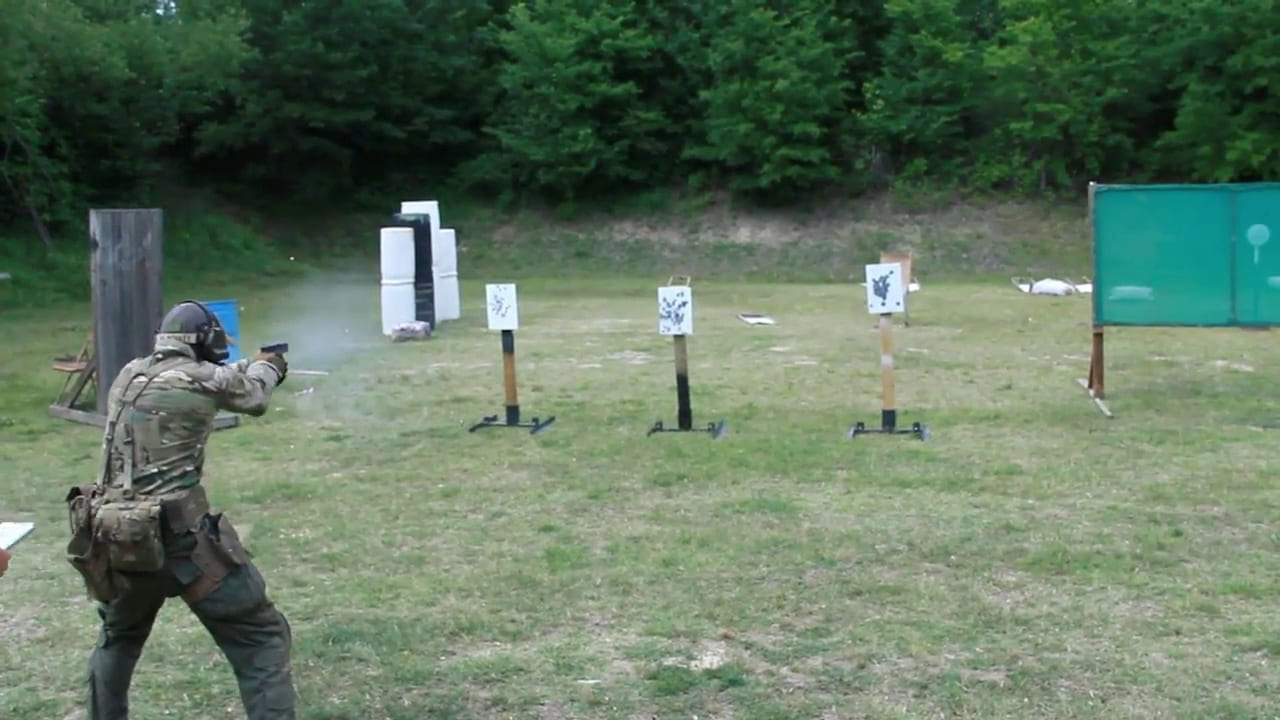}
		\caption{Original image.}
		\label{fig:darkDatasetExample_a}
	\end{subfigure}\hspace{12pt}
	\vspace{1pt}
	\begin{subfigure}[c]{0.48\textwidth}
		\includegraphics[width=\textwidth]{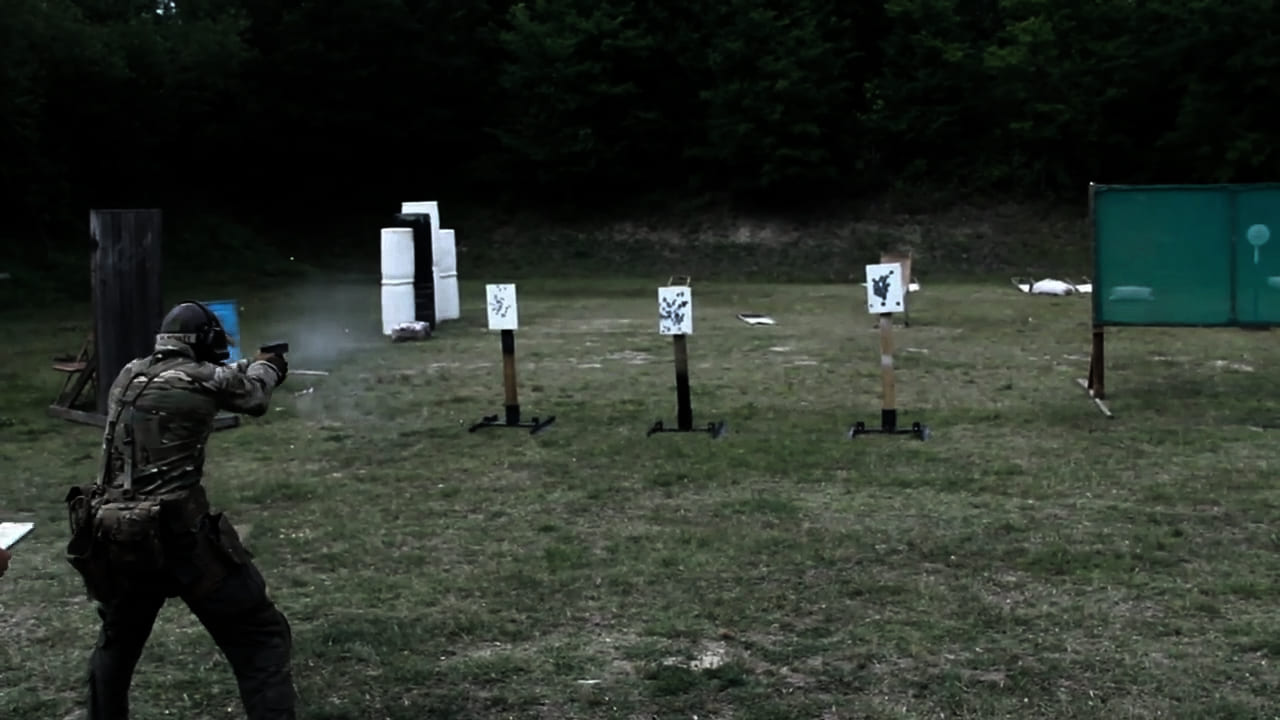}
		\caption{Darkened version.}
		\label{fig:darkDatasetExample_b}
	\end{subfigure}
	\caption{Test set A - Dark data.}
	\label{fig:darkDatasetExample}
\end{figure*}

A darkened version of the test set A was generated to simulate the performance of the trained models under poor illumination conditions. This scenario was obtained by modifying the \textit{Value} component in the HSV color space for all images in the test set A.~\autoref{fig:darkDatasetExample} shows an example image.

\begin{figure}[htbp]
	\centering
	\includegraphics[width=0.9\textwidth]{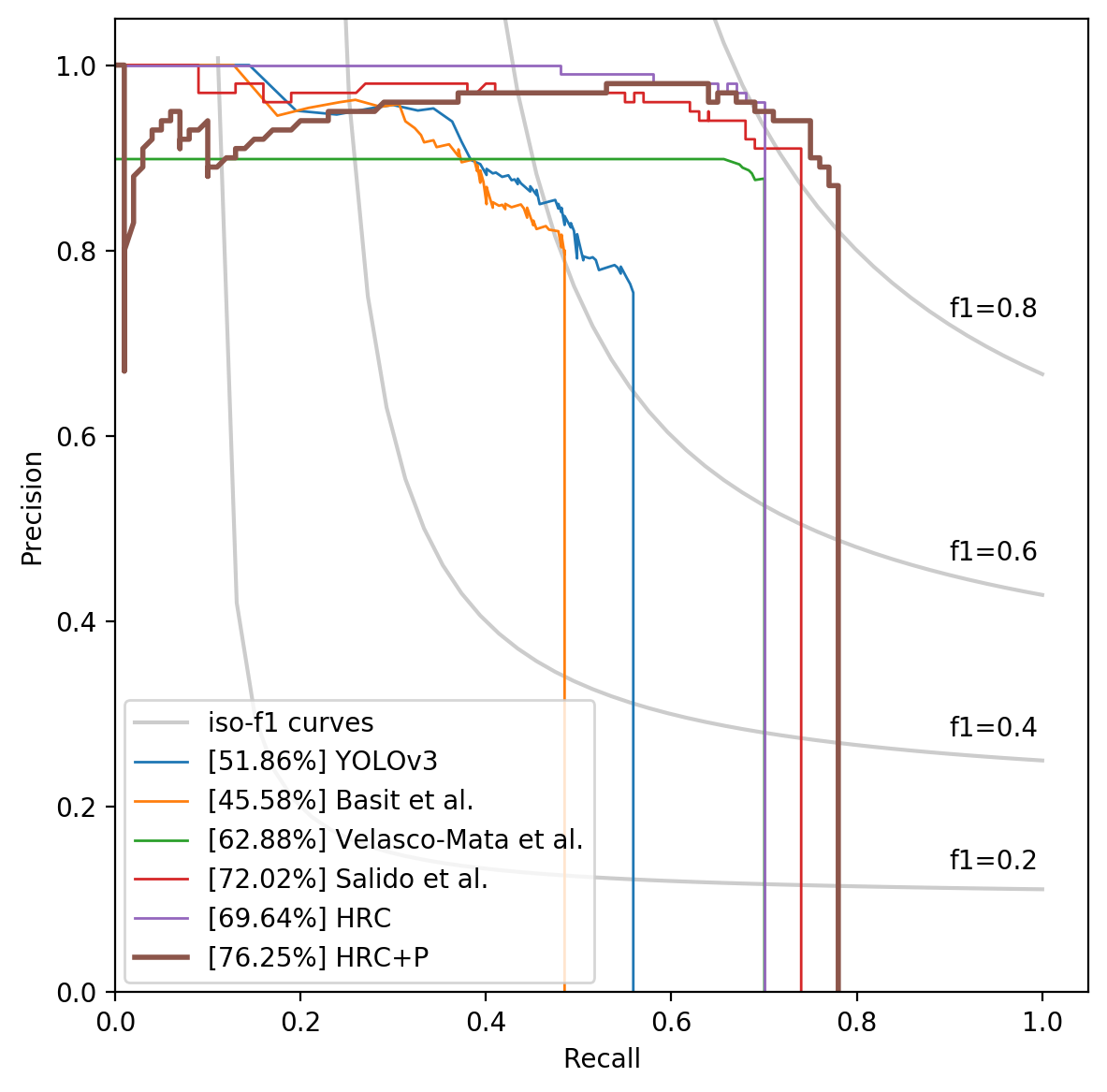}
	\caption{Precision-Recall curves obtained for test set A with the dark images.}
	\label{fig:apDark}
\end{figure}

The results obtained for these darkened images are summarized in~\autoref{tab:tableResults_dark}. Precision and Recall values are calculated with a 0.5 prediction score threshold and Precision-Recall curves of all methods are shown in~\autoref{fig:apDark}.
	

\begin{table*}[ht]
	\caption{Evaluation metrics for test set A with the dark images.}
	\label{tab:tableResults_dark}
	\vskip 0.1in
	\begin{center}
		\begin{tabular}{|c|c|c|c|}
			\cline{1-4}
			\textbf{Method}                                     & \textbf{Precision 0.5} & \textbf{Recall 0.5} & \textbf{AP}    \\  \cline{1-4}
			\textbf{YOLOv3~\cite{farhadi2018yolov3}}            & 0.8043                 & 0.4983              & 51.86          \\
			\textbf{Basit el al.~\cite{basit2020localizing}}    & 0.8561                 & 0.4007              & 45.58          \\
			\textbf{Velasco-Mata et al.~\cite{velasco2021human}}       & 0.8831                 & 0.6869              & 62.88          \\
			\textbf{Salido el al.~\cite{salido2021}}            & 0.9378                 & 0.6599              & 72.02          \\
			\textbf{HRC}                                        & \textbf{0.9751}        & 0.6599              & 69.64          \\
			\textbf{HRC+P}                                      & 0.8937                 & \textbf{0.7643}     & \textbf{76.25} \\
			\cline{1-4}
		\end{tabular}
	\end{center}
\end{table*}

Again, in these modified images the HRC+P method obtains the highest AP score, with more than 4\% of improvement over the second best, showing that the proposed pose-combined method can be useful in unfavorable lighting conditions.

\subsection{Test set A - Far data}

\begin{figure*}[htbp]
	\centering
	\begin{subfigure}[c]{0.48\textwidth}
		\includegraphics[width=\textwidth]{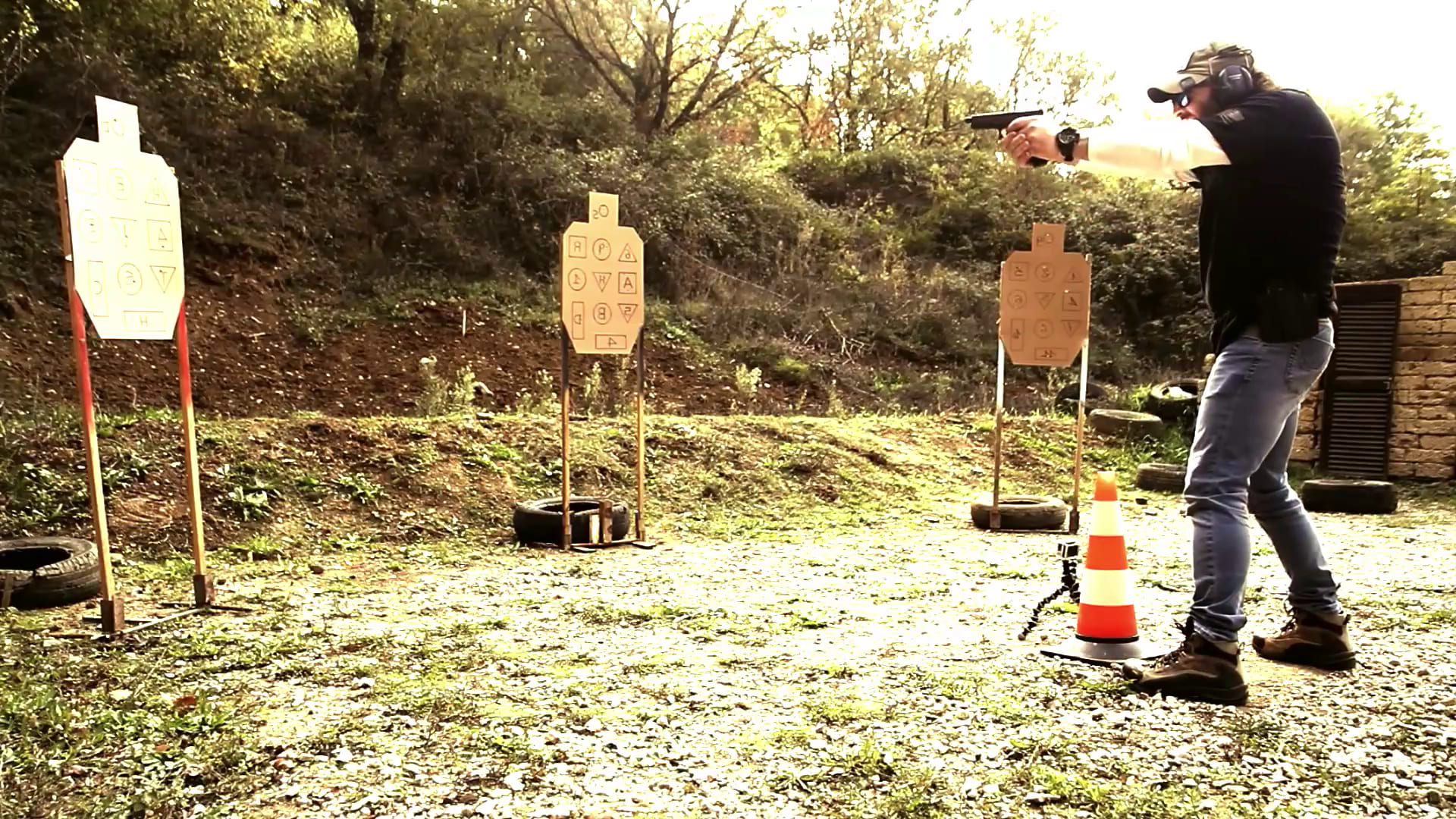}
		\caption{Original image.}
		\label{fig:farDatasetExample_a}
	\end{subfigure}\hspace{12pt}
	\vspace{1pt}
	\begin{subfigure}[c]{0.48\textwidth}
		\includegraphics[width=\textwidth]{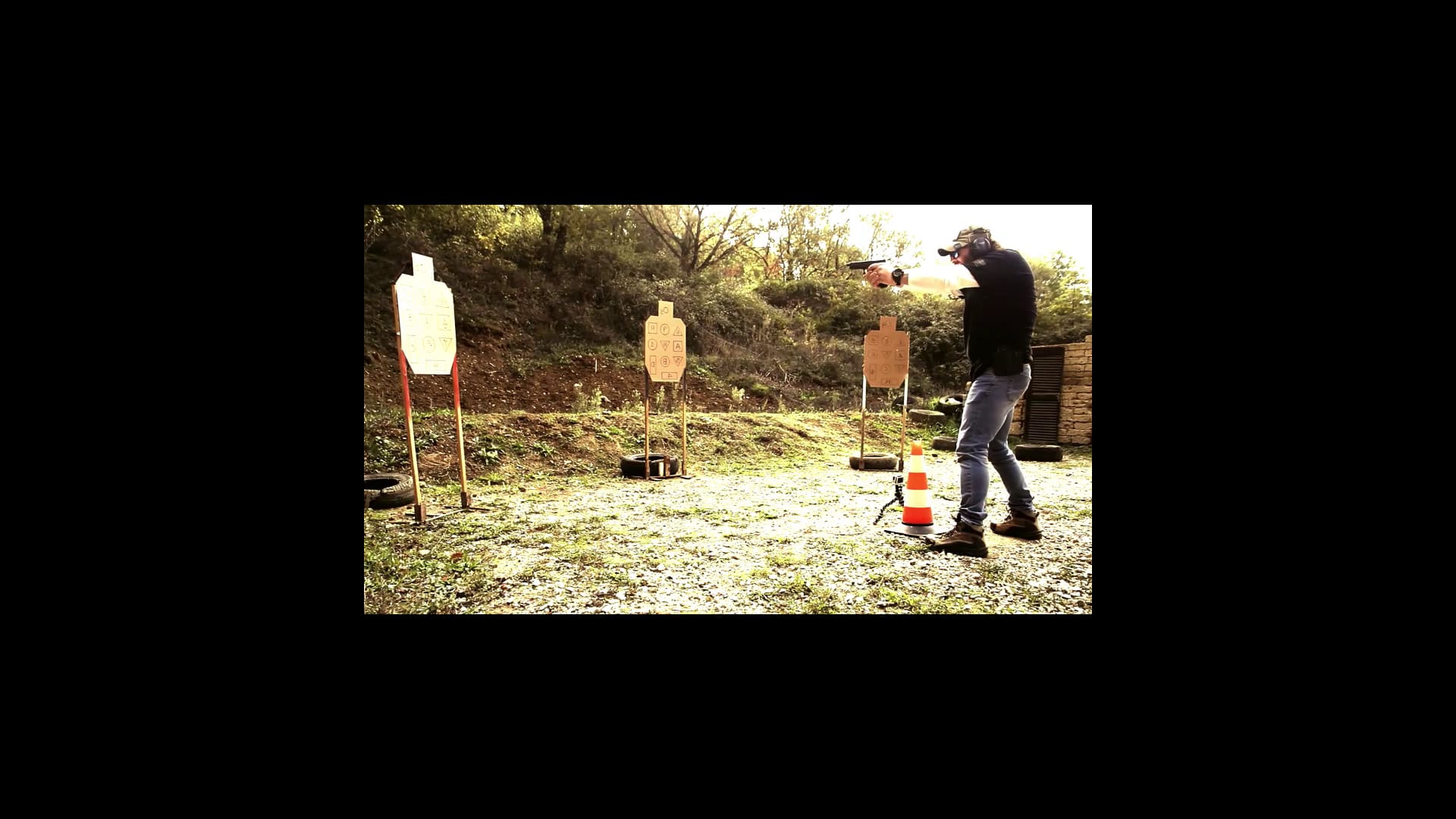}
		\caption{Far version.}
		\label{fig:farDatasetExample_b}
	\end{subfigure}
	\caption{Test set A - Far data.}
	\label{fig:farDatasetExample}
\end{figure*}

Camera distance can also be a relevant factor in detecting small objects such as handguns, especially in cases where only the visual appearance of the object is used for detection. To test this scenario, another version of the test set A has been generated, reducing image size by half and filling the rest of the image with black pixels. In~\autoref{fig:farDatasetExample}, an example of this transformation is presented.

\begin{figure}[!htb]
	\centering
	\includegraphics[width=0.9\textwidth]{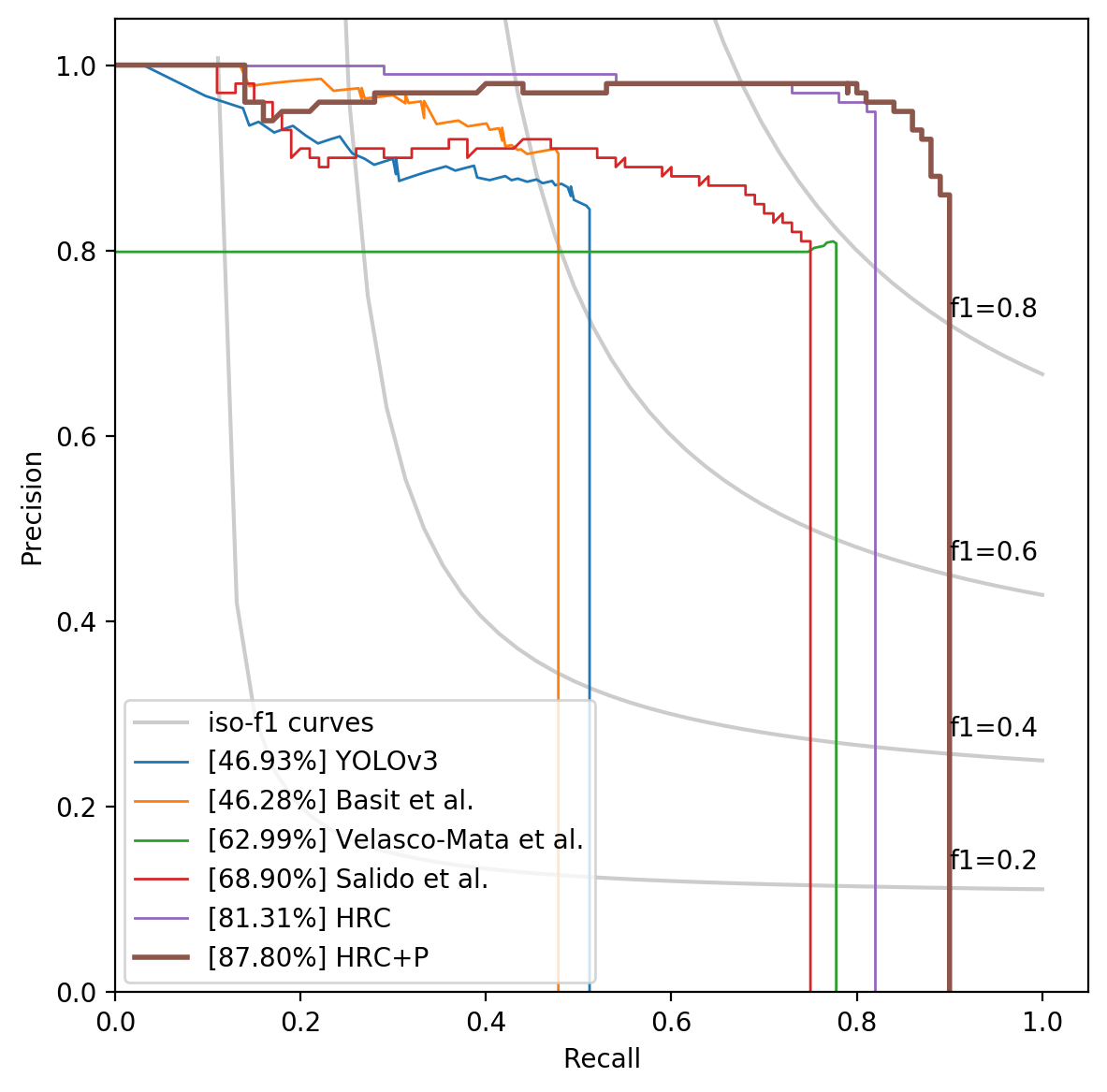}
	\caption{Precision-Recall curves obtained for test set A with the distance-simulated images.}
	\label{fig:apFar}
\end{figure}

The results obtained for these far images are summarized in~\autoref{tab:tableResults_far}. Precision and Recall values are calculated with a 0.5 prediction score threshold and Precision-Recall curves of all methods are shown in~\autoref{fig:apFar}.


\begin{table*}[ht]
\caption{Evaluation metrics for test set A with the distance-simulated images.}
\label{tab:tableResults_far}
\vskip 0.1in
	\begin{center}
		\begin{tabular}{|c|c|c|c|}
			\cline{1-4}
			\textbf{Method}                                     & \textbf{Precision 0.5} & \textbf{Recall 0.5} & \textbf{AP}    \\  \cline{1-4}
			\textbf{YOLOv3~\cite{farhadi2018yolov3}}            & 0.8766                 & 0.4545              & 46.93          \\
			\textbf{Basit el al.~\cite{basit2020localizing}}    & 0.9350                 & 0.3872              & 46.28          \\
			\textbf{Velasco-Mata et al.~\cite{velasco2021human}}       & 0.8092                 & 0.7710              & 62.99          \\
			\textbf{Salido el al.~\cite{salido2021}}            & 0.8449                 & 0.6970              & 68.90          \\
			\textbf{HRC}                                        & \textbf{0.9657}        & 0.7576              & 81.31          \\
			\textbf{HRC+P}                                      & 0.9123                 & \textbf{0.8754}     & \textbf{87.80} \\
			\cline{1-4}
		\end{tabular}
	\end{center}
\end{table*}

The HRC+P model also presents the highest AP score in this last test scenario, improving in 6\% the performance of the HRC method. The results show that this reduction in the size of the objects present in the image significantly affects methods that are based exclusively on the appearance of the object, such as YOLOv3. Pose-based methods show even higher metrics. This can be explained due to the fact that the pose estimation is not severely affected by this larger distance to the camera, being able to accurately detect the pose keypoints in the image.

\subsection{Test set B - Monash dataset}

\begin{figure}[!htb]
	\centering
	\includegraphics[width=0.9\textwidth]{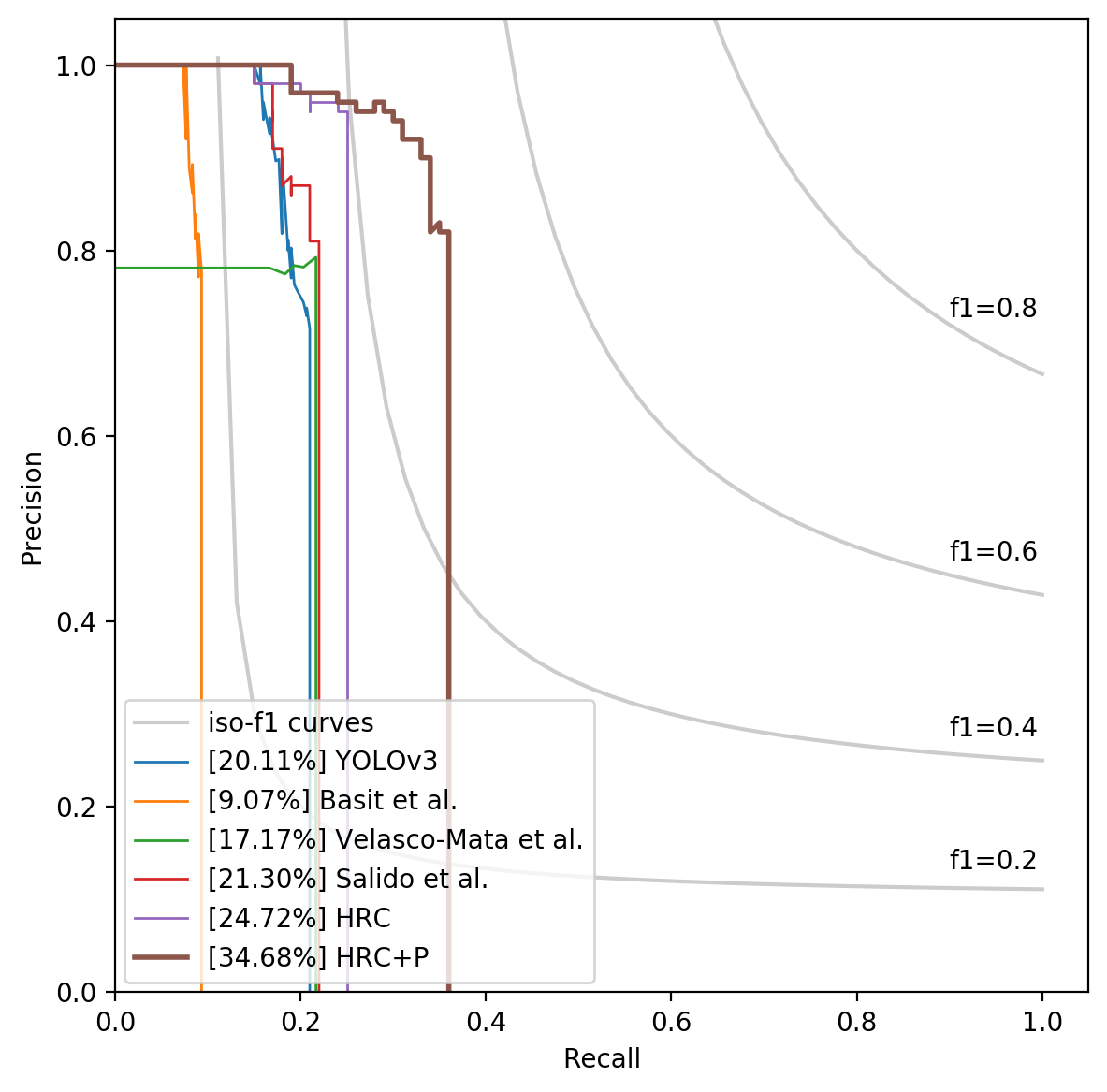}
	\caption{Precision-Recall curves obtained for test set B with the Monash images.}
	\label{fig:apMonash}
\end{figure}

For the last experiment, the realistic Monash Guns Dataset~\cite{lim97deep} has been used to test the methods compared in this work. This dataset shows people holding handguns in a variety of real-world CCTV surveillance environments. The results obtained for these test images are summarized in~\autoref{tab:tableResults_monash}. Precision and Recall values are calculated with a 0.5 prediction score threshold and Precision-Recall curves of all methods are shown in~\autoref{fig:apMonash}.

\begin{table*}[ht]
\caption{Evaluation metrics for test set B - Monash data}
\label{tab:tableResults_monash}
\vskip 0.1in
	\begin{center}
		\begin{tabular}{|c|c|c|c|}
			\cline{1-4}
			\textbf{Method}                                     & \textbf{Precision 0.5} & \textbf{Recall 0.5} & \textbf{AP}    \\  \cline{1-4}
			\textbf{YOLOv3~\cite{farhadi2018yolov3}}            & 0.8571                 & 0.1800              & 20.11          \\
			\textbf{Basit el al.~\cite{basit2020localizing}}    & 0.8929                 & 0.0833              &  9.07          \\
			\textbf{Velasco-Mata et al.~\cite{velasco2021human}}       & 0.7848                 & 0.2067              & 17.17          \\
			\textbf{Salido el al.~\cite{salido2021}}            & 0.8833                 & 0.1767              & 21.30          \\
			\textbf{HRC}                                        & \textbf{0.9683}        & 0.2033              & 24.72          \\
			\textbf{HRC+P}                                      & 0.9018                 & \textbf{0.3367}     & \textbf{34.68} \\
			\cline{1-4}
		\end{tabular}
	\end{center}
\end{table*}

For this test set, all methods show a significantly lower performance. This can be explained by the differences with respect to the training images in terms of lighting, camera perspective and image distortions. However, the proposed pose-combined method (HRC+P) still shows the highest AP score, improving the no-combined version (HRC) in approximately 10\%.

\subsection{Example images}

\begin{figure}[htb]
	\centering
	\begin{subfigure}[t]{0.49\textwidth}
		\includegraphics[width=\textwidth]{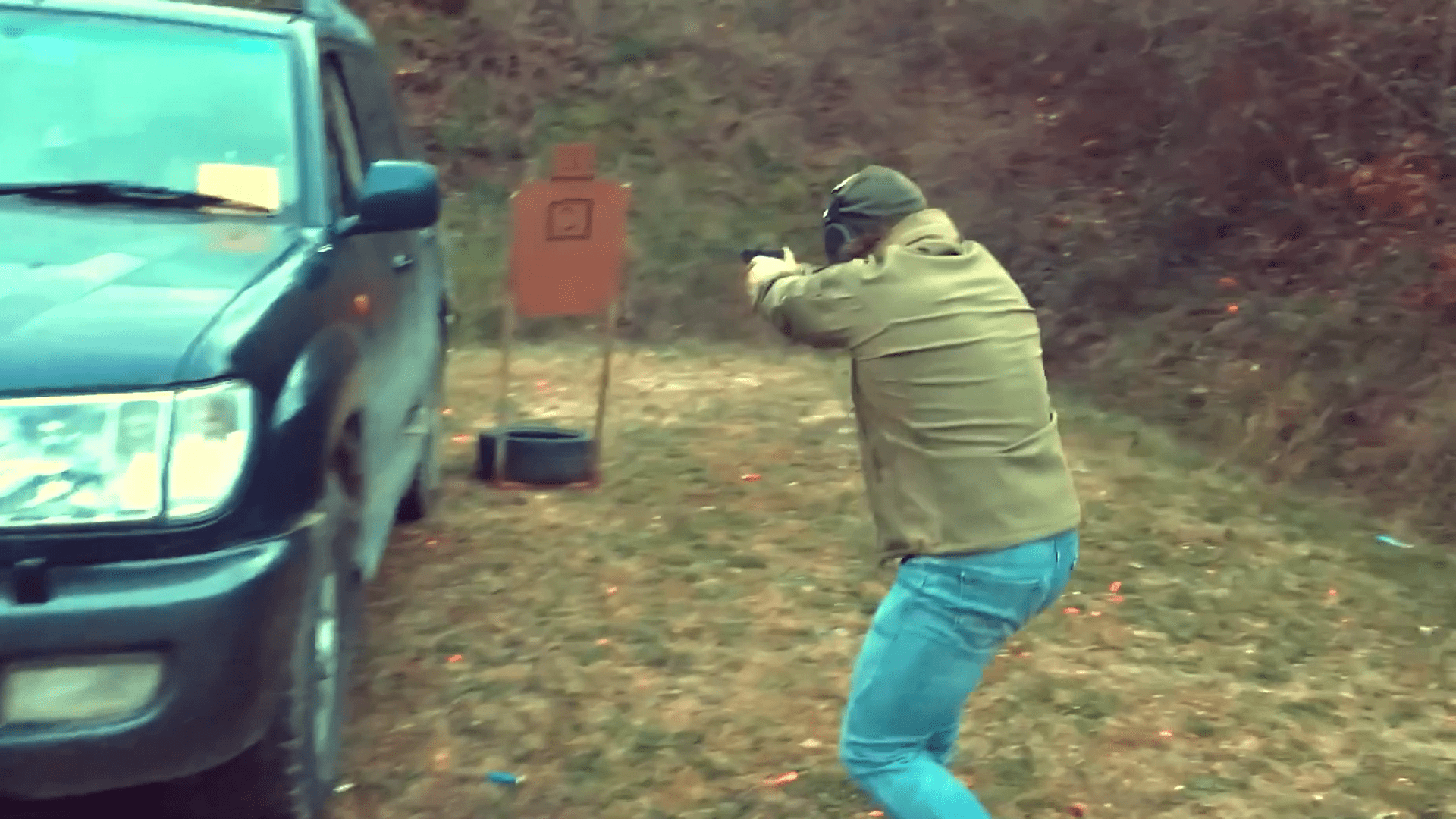}
		\caption{Example 1 - HRC}
		\label{fig:example_1_pgc_1}
	\end{subfigure}
	\begin{subfigure}[t]{0.49\textwidth}
		\includegraphics[width=\textwidth]{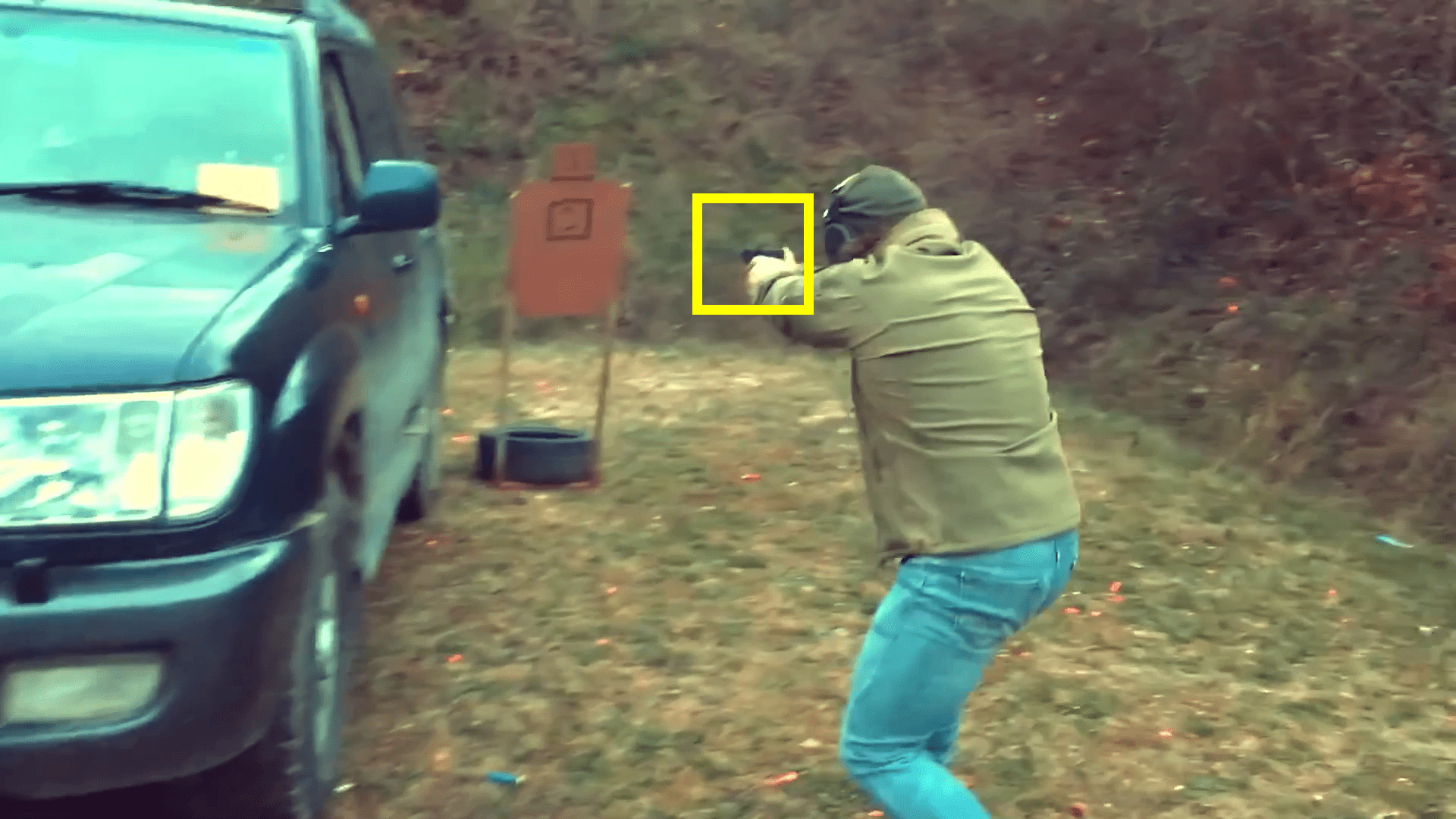}
		\caption{Example 1 - HRC+P}
		\label{fig:example_1_pgc_2}
	\end{subfigure}
	\begin{subfigure}[t]{0.49\textwidth}
		\includegraphics[width=\textwidth]{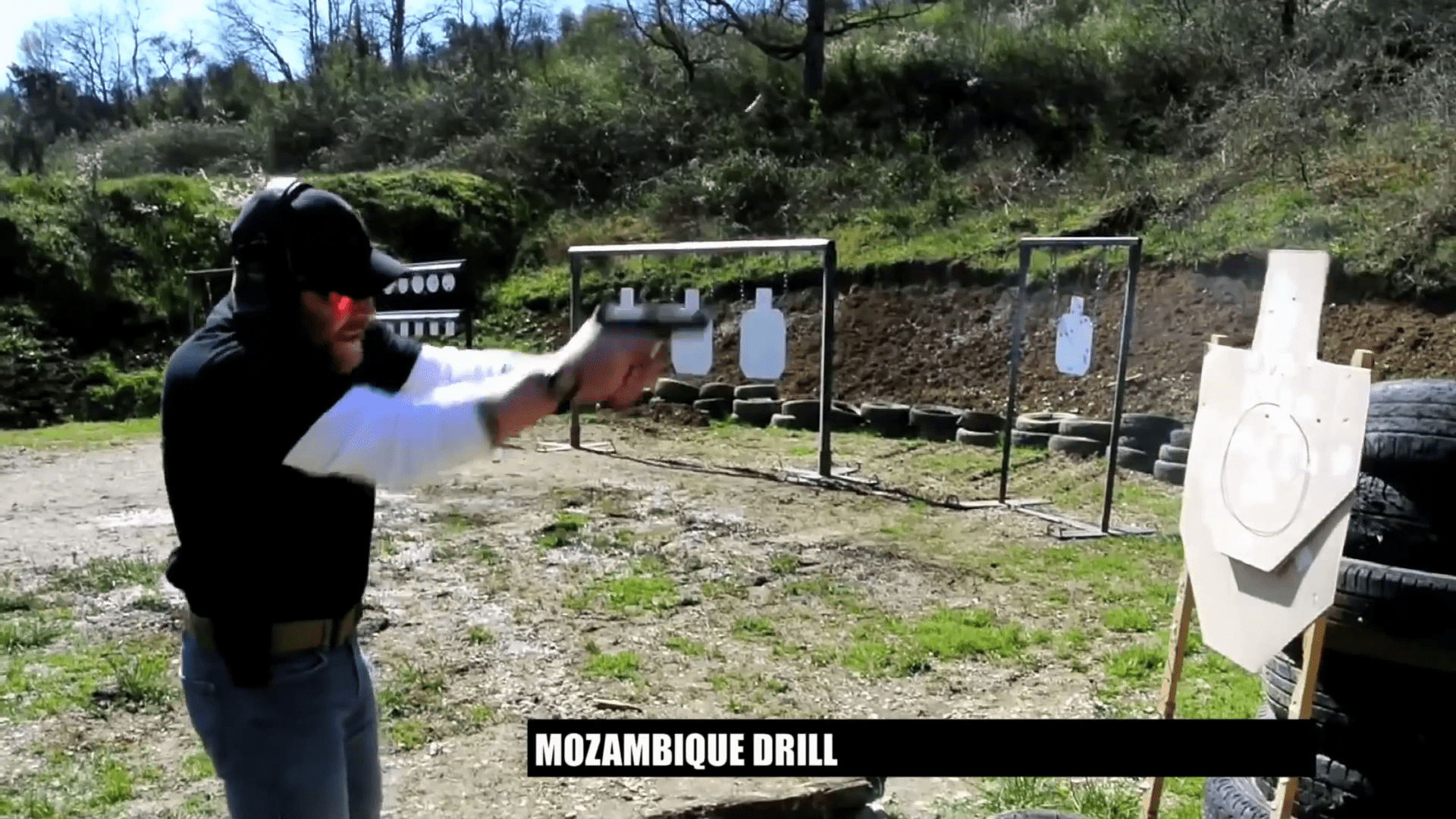}
		\caption{Example 2 - HRC}
		\label{fig:example_2_pgc_1}
	\end{subfigure}
	\begin{subfigure}[t]{0.49\textwidth}
		\includegraphics[width=\textwidth]{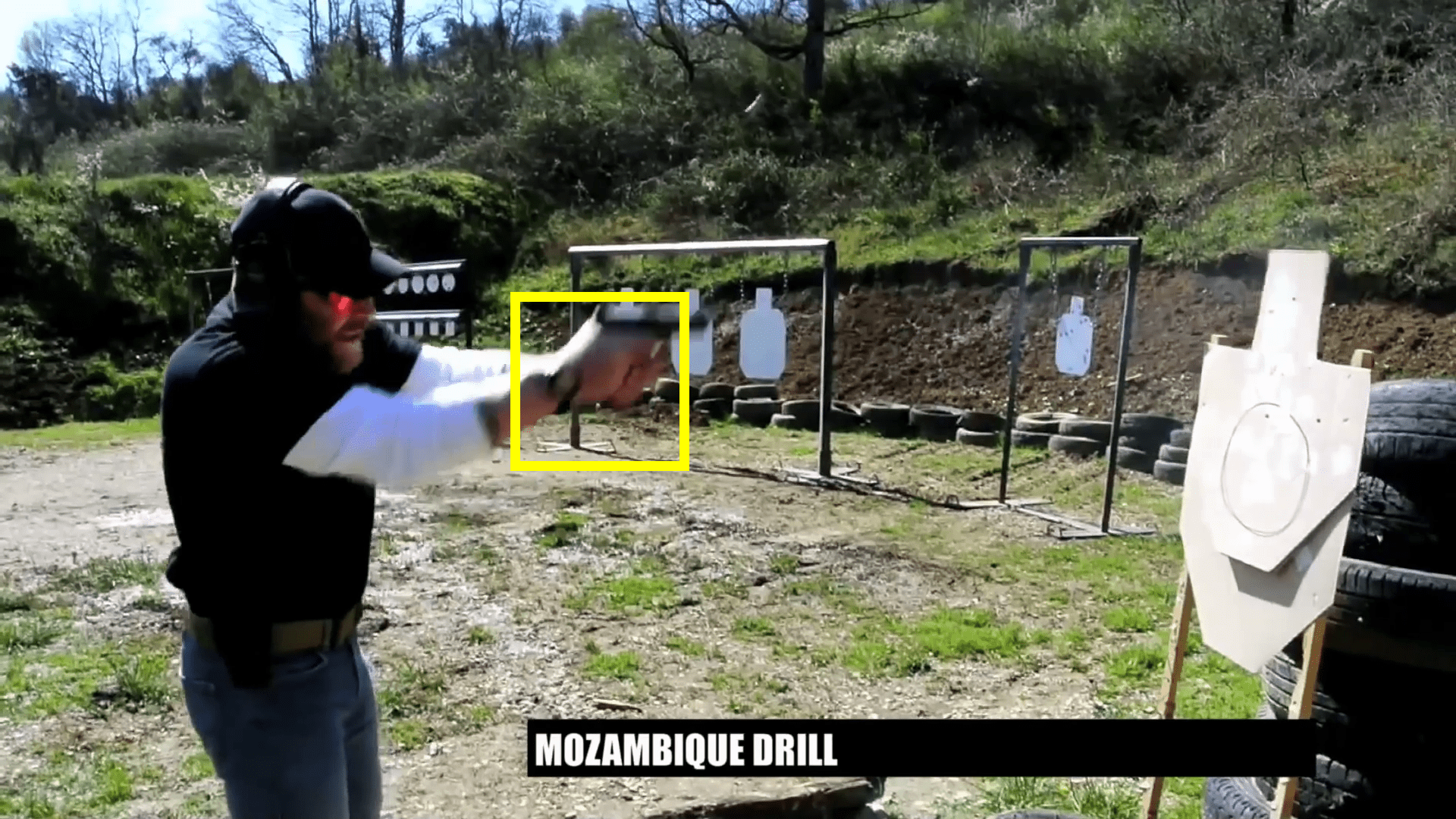}
		\caption{Example 2 - HRC+P}
		\label{fig:example_2_pgc_2}
	\end{subfigure}
	\caption{HRC and HRC+P detection examples 1}
	\label{fig:examples1_2}
\end{figure}

The two proposed approaches (HRC and HRC+P) present the best average performance in all studied scenarios. The pose-combined method (HRC+P) obtains better results in terms of Recall and AP. In~\autoref{fig:examples1_2} two example images are shown to illustrate these results. In both examples the handgun is not detected with the HRC method. In the first example (\autoref{fig:example_1_pgc_1} and~\autoref{fig:example_1_pgc_2}) the handgun is almost completely occluded and in the second example (\autoref{fig:example_2_pgc_1} and~\autoref{fig:example_2_pgc_2}) the handgun area is blurry. Conversely, the HRC+P method is able to locate both of them thanks to the clear shooting poses.

\begin{figure}[htb]
	\centering
	\begin{subfigure}[t]{0.49\textwidth}
		\includegraphics[width=\textwidth]{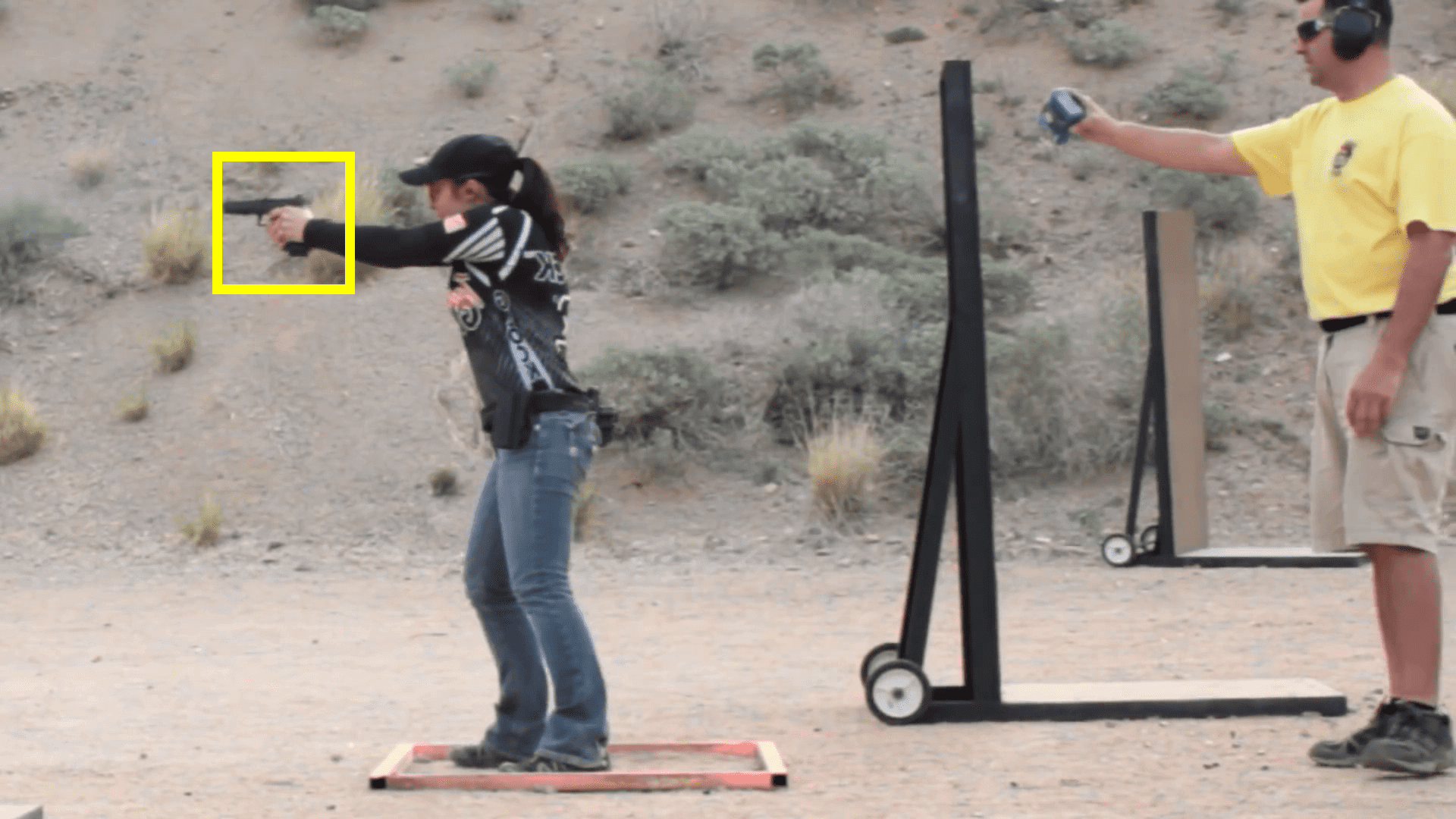}
		\caption{Example 3 - HRC}
		\label{fig:example_3_pgc_1}
	\end{subfigure}
	\begin{subfigure}[t]{0.49\textwidth}
		\includegraphics[width=\textwidth]{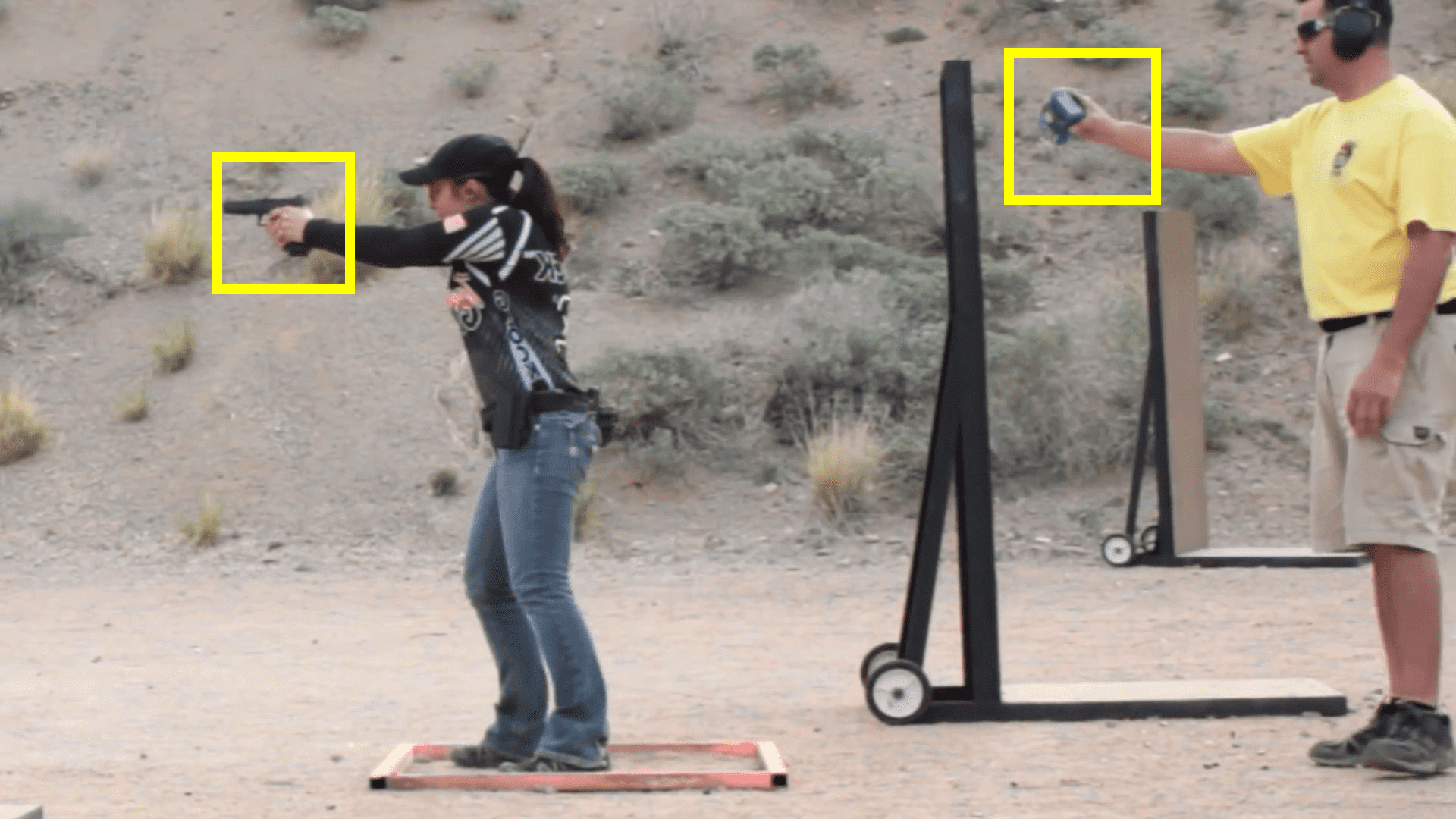}
		\caption{Example 3 - HRC+P}
		\label{fig:example_3_pgc_2}
	\end{subfigure}
	\caption{HRC and HRC+P detection examples 2}
	\label{fig:examples_3}
\end{figure}

In terms of Precision HRC obtains better results.~\autoref{fig:examples_3} shows an example of a false positive detection in the HRC+P method. In this case, both right hand object and body pose of the second subject have caused an incorrect detection. On the other hand, the HRC approach can classify all hand regions in the image correctly.

\section{Conclusions}
\label{S:6}

The 2D human pose is widely used in tasks such as action or gesture recognition. However, for the detection of threats or dangerous objects such as firearms, most of the proposed methods are based only on the visual appearance of the objects, without taking into account the human pose or another additional information.

In this work a novel method that combines in the same architecture the visual appearance of the handgun with the 2D human pose information is proposed. There are certain situations in which the object cannot be viewed correctly due to camera distance, poor lighting conditions or partial or total occlusion. In these situations, the human body pose helps in detecting the presence of handguns that would not be detected without this additional information. On the other hand, as the pose information is used to classify only the hand regions of the people detected, it is possible to remove false positives that may appear in other locations of the image.

The tests performed with the different datasets show that the proposed method using the pose combination obtains better results in all cases. Especially interesting is the fact that metrics in the reduced size images are even higher than in the case of the original size images.

Automatic and real time handgun detection in CCTV video surveillance images is still an open problem and there is room for improvement. The authors hope that the proposed work can be used as inspiration for new approaches based on 2D human pose information to improve the overall detector performance in this kind of application. 

Finally, note that in real scenarios common hand-held objects such as cell phones, keys or wallets may be an important source of false positives or misclassifications. In future work this aspect will be addressed with more specific methods.

\section*{Acknowledgment}
\label{S:8}

This work was partially funded by the Spanish Ministry of Economy and Business [grant number TIN2017-82113-C2-2-R], the Autonomous Government of Castilla-La Mancha [grant number SBPLY/17/180501/000543] and the ERDF. Author J. Ruiz-Santaquiteria was supported by Postgraduate Grant from the Spanish Ministry of Science, Innovation, and Universities [grant number PRE2018-083772].

\bibliographystyle{elsarticle-num-names.bst}
\bibliography{Bibliography.bib}

\begin{thebibliography}{35}
\expandafter\ifx\csname natexlab\endcsname\relax\def\natexlab#1{#1}\fi
\providecommand{\url}[1]{\texttt{#1}}
\providecommand{\href}[2]{#2}
\providecommand{\path}[1]{#1}
\providecommand{\DOIprefix}{doi:}
\providecommand{\ArXivprefix}{arXiv:}
\providecommand{\URLprefix}{URL: }
\providecommand{\Pubmedprefix}{pmid:}
\providecommand{\doi}[1]{\href{http://dx.doi.org/#1}{\path{#1}}}
\providecommand{\Pubmed}[1]{\href{pmid:#1}{\path{#1}}}
\providecommand{\bibinfo}[2]{#2}
\ifx\xfnm\relax \def\xfnm[#1]{\unskip,\space#1}\fi
\bibitem[{Ainsworth(2002)}]{ainsworth2002buyer}
\bibinfo{author}{T.~Ainsworth},
\newblock \bibinfo{title}{Buyer beware},
\newblock \bibinfo{journal}{Security Oz} \bibinfo{volume}{19}
  (\bibinfo{year}{2002}) \bibinfo{pages}{18--26}.
\bibitem[{Velastin et~al.(2006)Velastin, Boghossian, and
  Vicencio-Silva}]{velastin2006motion}
\bibinfo{author}{S.~A. Velastin}, \bibinfo{author}{B.~A. Boghossian},
  \bibinfo{author}{M.~A. Vicencio-Silva},
\newblock \bibinfo{title}{A motion-based image processing system for detecting
  potentially dangerous situations in underground railway stations},
\newblock \bibinfo{journal}{Transportation Research Part C: Emerging
  Technologies} \bibinfo{volume}{14} (\bibinfo{year}{2006})
  \bibinfo{pages}{96--113}.
\bibitem[{Enr{\'\i}quez et~al.(2019)Enr{\'\i}quez, Soria,
  {\'A}lvarez-Garc{\'\i}a, Caparrini, Velasco, Deniz, and
  Vallez}]{enriquez2019vision}
\bibinfo{author}{F.~Enr{\'\i}quez}, \bibinfo{author}{L.~M. Soria},
  \bibinfo{author}{J.~A. {\'A}lvarez-Garc{\'\i}a}, \bibinfo{author}{F.~S.
  Caparrini}, \bibinfo{author}{F.~Velasco}, \bibinfo{author}{O.~Deniz},
  \bibinfo{author}{N.~Vallez},
\newblock \bibinfo{title}{Vision and crowdsensing technology for an optimal
  response in physical-security},
\newblock in: \bibinfo{booktitle}{International Conference on Computational
  Science}, \bibinfo{organization}{Springer}, \bibinfo{year}{2019}, pp.
  \bibinfo{pages}{15--26}.
\bibitem[{{Everytown for Gun Safety}(2020)}]{gunfireUSA}
\bibinfo{author}{{Everytown for Gun Safety}}, \bibinfo{title}{Gunfire on
  {School} {Grounds} in the {United} {States}},
  \bibinfo{howpublished}{\url{https://everytownresearch.org/gunfire-in-school/\#ns}},
  \bibinfo{year}{2020}. \bibinfo{note}{Accessed: 20/07/2021}.
\bibitem[{Tessler et~al.(2017)Tessler, Mooney, Witt, O’Connell, Jenness,
  Vavilala, and Rivara}]{tessler2017use}
\bibinfo{author}{R.~A. Tessler}, \bibinfo{author}{S.~J. Mooney},
  \bibinfo{author}{C.~E. Witt}, \bibinfo{author}{K.~O’Connell},
  \bibinfo{author}{J.~Jenness}, \bibinfo{author}{M.~S. Vavilala},
  \bibinfo{author}{F.~P. Rivara},
\newblock \bibinfo{title}{Use of firearms in terrorist attacks: differences
  between the {United} {States}, {Canada}, {Europe}, {Australia}, and {New}
  {Zealand}},
\newblock \bibinfo{journal}{JAMA internal medicine} \bibinfo{volume}{177}
  (\bibinfo{year}{2017}) \bibinfo{pages}{1865--1868}.
\bibitem[{Vallez et~al.(2013)Vallez, Bueno, and Deniz}]{vallez2013false}
\bibinfo{author}{N.~Vallez}, \bibinfo{author}{G.~Bueno},
  \bibinfo{author}{O.~Deniz},
\newblock \bibinfo{title}{False positive reduction in detector implantation},
\newblock in: \bibinfo{booktitle}{Conference on Artificial Intelligence in
  Medicine in Europe}, \bibinfo{organization}{Springer}, \bibinfo{year}{2013},
  pp. \bibinfo{pages}{181--185}.
\bibitem[{Castillo et~al.(2019)Castillo, Tabik, P{\'e}rez, Olmos, and
  Herrera}]{castillo2019brightness}
\bibinfo{author}{A.~Castillo}, \bibinfo{author}{S.~Tabik},
  \bibinfo{author}{F.~P{\'e}rez}, \bibinfo{author}{R.~Olmos},
  \bibinfo{author}{F.~Herrera},
\newblock \bibinfo{title}{Brightness guided preprocessing for automatic cold
  steel weapon detection in surveillance videos with deep learning},
\newblock \bibinfo{journal}{Neurocomputing} \bibinfo{volume}{330}
  (\bibinfo{year}{2019}) \bibinfo{pages}{151--161}.
\bibitem[{Nercessian et~al.(2008)Nercessian, Panetta, and
  Agaian}]{nercessian2008automatic}
\bibinfo{author}{S.~Nercessian}, \bibinfo{author}{K.~Panetta},
  \bibinfo{author}{S.~Agaian},
\newblock \bibinfo{title}{Automatic detection of potential threat objects in
  x-ray luggage scan images},
\newblock in: \bibinfo{booktitle}{2008 IEEE Conference on Technologies for
  Homeland Security}, \bibinfo{organization}{IEEE}, \bibinfo{year}{2008}, pp.
  \bibinfo{pages}{504--509}.
\bibitem[{Xiao et~al.(2015)Xiao, Lu, Yan, Wu, and Ren}]{xiao2015automatic}
\bibinfo{author}{Z.~Xiao}, \bibinfo{author}{X.~Lu}, \bibinfo{author}{J.~Yan},
  \bibinfo{author}{L.~Wu}, \bibinfo{author}{L.~Ren},
\newblock \bibinfo{title}{Automatic detection of concealed pistols using
  passive millimeter wave imaging},
\newblock in: \bibinfo{booktitle}{2015 IEEE International Conference on Imaging
  Systems and Techniques (IST)}, \bibinfo{organization}{IEEE},
  \bibinfo{year}{2015}, pp. \bibinfo{pages}{1--4}.
\bibitem[{Flitton et~al.(2013)Flitton, Breckon, and
  Megherbi}]{flitton2013comparison}
\bibinfo{author}{G.~Flitton}, \bibinfo{author}{T.~P. Breckon},
  \bibinfo{author}{N.~Megherbi},
\newblock \bibinfo{title}{A comparison of {3D} interest point descriptors with
  application to airport baggage object detection in complex {CT} imagery},
\newblock \bibinfo{journal}{Pattern Recognition} \bibinfo{volume}{46}
  (\bibinfo{year}{2013}) \bibinfo{pages}{2420--2436}.
\bibitem[{Flitton et~al.(2015)Flitton, Mouton, and Breckon}]{flitton2015object}
\bibinfo{author}{G.~Flitton}, \bibinfo{author}{A.~Mouton},
  \bibinfo{author}{T.~P. Breckon},
\newblock \bibinfo{title}{Object classification in {3D} baggage security
  computed tomography imagery using visual codebooks},
\newblock \bibinfo{journal}{Pattern Recognition} \bibinfo{volume}{48}
  (\bibinfo{year}{2015}) \bibinfo{pages}{2489--2499}.
\bibitem[{Tiwari and Verma(2015)}]{tiwari2015computer}
\bibinfo{author}{R.~K. Tiwari}, \bibinfo{author}{G.~K. Verma},
\newblock \bibinfo{title}{A computer vision based framework for visual gun
  detection using {Harris} interest point detector},
\newblock \bibinfo{journal}{Procedia Computer Science} \bibinfo{volume}{54}
  (\bibinfo{year}{2015}) \bibinfo{pages}{703--712}.
\bibitem[{Halima and Hosam(2016)}]{halima2016bag}
\bibinfo{author}{N.~B. Halima}, \bibinfo{author}{O.~Hosam},
\newblock \bibinfo{title}{Bag of words based surveillance system using support
  vector machines},
\newblock \bibinfo{journal}{Int. J. Secur. Appl} \bibinfo{volume}{10}
  (\bibinfo{year}{2016}) \bibinfo{pages}{331--346}.
\bibitem[{Grega et~al.(2016)Grega, Matiola{\'n}ski, Guzik, and
  Leszczuk}]{grega2016automated}
\bibinfo{author}{M.~Grega}, \bibinfo{author}{A.~Matiola{\'n}ski},
  \bibinfo{author}{P.~Guzik}, \bibinfo{author}{M.~Leszczuk},
\newblock \bibinfo{title}{Automated detection of firearms and knives in a
  {CCTV} image},
\newblock \bibinfo{journal}{Sensors} \bibinfo{volume}{16}
  (\bibinfo{year}{2016}) \bibinfo{pages}{47}.
\bibitem[{Gelana and Yadav(2019)}]{gelana2019firearm}
\bibinfo{author}{F.~Gelana}, \bibinfo{author}{A.~Yadav},
\newblock \bibinfo{title}{Firearm detection from surveillance cameras using
  image processing and machine learning techniques},
\newblock in: \bibinfo{booktitle}{Smart Innovations in Communication and
  Computational Sciences}, \bibinfo{publisher}{Springer}, \bibinfo{year}{2019},
  pp. \bibinfo{pages}{25--34}.
\bibitem[{Girshick(2015)}]{girshick2015fast}
\bibinfo{author}{R.~Girshick},
\newblock \bibinfo{title}{Fast {R-CNN}},
\newblock in: \bibinfo{booktitle}{Proceedings of the IEEE international
  conference on computer vision}, \bibinfo{year}{2015}, pp.
  \bibinfo{pages}{1440--1448}.
\bibitem[{Ren et~al.(2015)Ren, He, Girshick, and Sun}]{ren2015faster}
\bibinfo{author}{S.~Ren}, \bibinfo{author}{K.~He},
  \bibinfo{author}{R.~Girshick}, \bibinfo{author}{J.~Sun},
\newblock \bibinfo{title}{Faster {R-CNN}: Towards real-time object detection
  with region proposal networks},
\newblock \bibinfo{journal}{Advances in Neural Information Processing Systems}
  (\bibinfo{year}{2015}) \bibinfo{pages}{91--99}.
\bibitem[{Verma and Dhillon(2017)}]{verma2017handheld}
\bibinfo{author}{G.~K. Verma}, \bibinfo{author}{A.~Dhillon},
\newblock \bibinfo{title}{A handheld gun detection using faster {R-CNN} deep
  learning},
\newblock in: \bibinfo{booktitle}{Proceedings of the 7th International
  Conference on Computer and Communication Technology}, \bibinfo{year}{2017},
  pp. \bibinfo{pages}{84--88}.
\bibitem[{{{IMFDB}: {Internet} {Movie} {Firearms} {Database}}(2020)}]{imfdb}
\bibinfo{author}{{{IMFDB}: {Internet} {Movie} {Firearms} {Database}}},
  \bibinfo{howpublished}{\url{http://www.imfdb.org/wiki/Main_Page}},
  \bibinfo{year}{2020}. \bibinfo{note}{Accessed: 20/07/2021}.
\bibitem[{Olmos et~al.(2018)Olmos, Tabik, and Herrera}]{olmos2018automatic}
\bibinfo{author}{R.~Olmos}, \bibinfo{author}{S.~Tabik},
  \bibinfo{author}{F.~Herrera},
\newblock \bibinfo{title}{Automatic handgun detection alarm in videos using
  deep learning},
\newblock \bibinfo{journal}{Neurocomputing} \bibinfo{volume}{275}
  (\bibinfo{year}{2018}) \bibinfo{pages}{66--72}.
\bibitem[{Redmon et~al.(2016)Redmon, Divvala, Girshick, and
  Farhadi}]{redmon2016you}
\bibinfo{author}{J.~Redmon}, \bibinfo{author}{S.~Divvala},
  \bibinfo{author}{R.~Girshick}, \bibinfo{author}{A.~Farhadi},
\newblock \bibinfo{title}{You only look once: Unified, real-time object
  detection},
\newblock in: \bibinfo{booktitle}{Proceedings of the IEEE conference on
  computer vision and pattern recognition}, \bibinfo{year}{2016}, pp.
  \bibinfo{pages}{779--788}.
\bibitem[{Redmon and Farhadi(2017)}]{redmon2017yolo9000}
\bibinfo{author}{J.~Redmon}, \bibinfo{author}{A.~Farhadi},
\newblock \bibinfo{title}{Yolo9000: better, faster, stronger},
\newblock in: \bibinfo{booktitle}{Proceedings of the IEEE conference on
  computer vision and pattern recognition}, \bibinfo{year}{2017}, pp.
  \bibinfo{pages}{7263--7271}.
\bibitem[{Farhadi and Redmon(2018)}]{farhadi2018yolov3}
\bibinfo{author}{A.~Farhadi}, \bibinfo{author}{J.~Redmon},
\newblock \bibinfo{title}{Yolov3: An incremental improvement},
\newblock \bibinfo{journal}{Computer Vision and Pattern Recognition}
  (\bibinfo{year}{2018}).
\bibitem[{Warsi et~al.(2019)Warsi, Abdullah, Husen, Yahya, Khan, and
  Jawaid}]{warsi2019gun}
\bibinfo{author}{A.~Warsi}, \bibinfo{author}{M.~Abdullah},
  \bibinfo{author}{M.~N. Husen}, \bibinfo{author}{M.~Yahya},
  \bibinfo{author}{S.~Khan}, \bibinfo{author}{N.~Jawaid},
\newblock \bibinfo{title}{Gun detection system using yolov3},
\newblock in: \bibinfo{booktitle}{2019 IEEE International Conference on Smart
  Instrumentation, Measurement and Application (ICSIMA)},
  \bibinfo{organization}{IEEE}, \bibinfo{year}{2019}, pp.
  \bibinfo{pages}{1--4}.
\bibitem[{de~Azevedo~Kanehisa and de~Almeida~Neto(2019)}]{de2019firearm}
\bibinfo{author}{R.~F. de~Azevedo~Kanehisa},
  \bibinfo{author}{A.~de~Almeida~Neto},
\newblock \bibinfo{title}{Firearm detection using convolutional neural
  networks.},
\newblock in: \bibinfo{booktitle}{ICAART (2)}, \bibinfo{year}{2019}, pp.
  \bibinfo{pages}{707--714}.
\bibitem[{Abruzzo et~al.(2019)Abruzzo, Carey, Lowrance, Sturzinger, Arnold, and
  Korpela}]{abruzzo2019cascaded}
\bibinfo{author}{B.~Abruzzo}, \bibinfo{author}{K.~Carey},
  \bibinfo{author}{C.~Lowrance}, \bibinfo{author}{E.~Sturzinger},
  \bibinfo{author}{R.~Arnold}, \bibinfo{author}{C.~Korpela},
\newblock \bibinfo{title}{Cascaded neural networks for identification and
  posture-based threat assessment of armed people},
\newblock in: \bibinfo{booktitle}{2019 IEEE International Symposium on
  Technologies for Homeland Security (HST)}, \bibinfo{organization}{IEEE},
  \bibinfo{year}{2019}, pp. \bibinfo{pages}{1--7}.
\bibitem[{Basit et~al.(2020)Basit, Munir, Ali, Werghi, and
  Mahmood}]{basit2020localizing}
\bibinfo{author}{A.~Basit}, \bibinfo{author}{M.~A. Munir},
  \bibinfo{author}{M.~Ali}, \bibinfo{author}{N.~Werghi},
  \bibinfo{author}{A.~Mahmood},
\newblock \bibinfo{title}{Localizing firearm carriers by identifying
  human-object pairs},
\newblock in: \bibinfo{booktitle}{2020 IEEE International Conference on Image
  Processing (ICIP)}, \bibinfo{organization}{IEEE}, \bibinfo{year}{2020}, pp.
  \bibinfo{pages}{2031--2035}.
\bibitem[{Salido et~al.(2021)Salido, Lomas, Ruiz-Santaquiteria, and
  Deniz}]{salido2021}
\bibinfo{author}{J.~Salido}, \bibinfo{author}{V.~Lomas},
  \bibinfo{author}{J.~Ruiz-Santaquiteria}, \bibinfo{author}{O.~Deniz},
\newblock \bibinfo{title}{Automatic handgun detection with deep learning in
  video surveillance images},
\newblock \bibinfo{journal}{Applied Sciences} \bibinfo{volume}{11}
  (\bibinfo{year}{2021}). \URLprefix
  \url{https://www.mdpi.com/2076-3417/11/13/6085}.
  \DOIprefix\doi{10.3390/app11136085}.
\bibitem[{Velasco-Mata et~al.(2021)Velasco-Mata, Ruiz-Santaquiteria, Vallez,
  and Deniz}]{velasco2021human}
\bibinfo{author}{A.~Velasco-Mata}, \bibinfo{author}{J.~Ruiz-Santaquiteria},
  \bibinfo{author}{N.~Vallez}, \bibinfo{author}{O.~Deniz},
\newblock \bibinfo{title}{Using human pose information for handgun detection},
\newblock \bibinfo{journal}{Neural Computing and Applications}
  (\bibinfo{year}{2021}).
\bibitem[{Salazar~Gonz{\'{a}}lez et~al.(2020)Salazar~Gonz{\'{a}}lez, Zaccaro,
  {\'{A}}lvarez-Garc{\'{i}}a, Soria-Morillo, and
  Sancho~Caparrini}]{SalazarGonzalez2020}
\bibinfo{author}{J.~L. Salazar~Gonz{\'{a}}lez}, \bibinfo{author}{C.~Zaccaro},
  \bibinfo{author}{J.~A. {\'{A}}lvarez-Garc{\'{i}}a}, \bibinfo{author}{L.~M.
  Soria-Morillo}, \bibinfo{author}{F.~Sancho~Caparrini},
\newblock \bibinfo{title}{Real-time gun detection in cctv: An open problem},
\newblock \bibinfo{journal}{Neural Networks} \bibinfo{volume}{132}
  (\bibinfo{year}{2020}) \bibinfo{pages}{297 -- 308}.
\bibitem[{Lim et~al.(2021)Lim, Al~Jobayer, Baskaran, Lim, See, and
  Wong}]{lim97deep}
\bibinfo{author}{J.~Lim}, \bibinfo{author}{M.~I. Al~Jobayer},
  \bibinfo{author}{V.~M. Baskaran}, \bibinfo{author}{J.~M. Lim},
  \bibinfo{author}{J.~See}, \bibinfo{author}{K.~Wong},
\newblock \bibinfo{title}{Deep multi-level feature pyramids: Application for
  non-canonical firearm detection in video surveillance},
\newblock \bibinfo{journal}{Engineering Applications of Artificial
  Intelligence} \bibinfo{volume}{97} (\bibinfo{year}{2021})
  \bibinfo{pages}{104094}.
\bibitem[{Lin et~al.(2014)Lin, Maire, Belongie, Hays, Perona, Ramanan,
  Doll{\'a}r, and Zitnick}]{coco}
\bibinfo{author}{T.-Y. Lin}, \bibinfo{author}{M.~Maire},
  \bibinfo{author}{S.~Belongie}, \bibinfo{author}{J.~Hays},
  \bibinfo{author}{P.~Perona}, \bibinfo{author}{D.~Ramanan},
  \bibinfo{author}{P.~Doll{\'a}r}, \bibinfo{author}{C.~L. Zitnick},
\newblock \bibinfo{title}{Microsoft {COCO}: Common {Objects} in {Context}},
\newblock in: \bibinfo{booktitle}{European conference on computer vision},
  \bibinfo{organization}{Springer}, \bibinfo{year}{2014}, pp.
  \bibinfo{pages}{740--755}.
\bibitem[{Deng et~al.(2009)Deng, Dong, Socher, Li, Li, and Fei-Fei}]{imageNet}
\bibinfo{author}{J.~Deng}, \bibinfo{author}{W.~Dong},
  \bibinfo{author}{R.~Socher}, \bibinfo{author}{L.-J. Li},
  \bibinfo{author}{K.~Li}, \bibinfo{author}{L.~Fei-Fei},
\newblock \bibinfo{title}{{ImageNet: A Large-Scale Hierarchical Image
  Database}},
\newblock in: \bibinfo{booktitle}{CVPR09}, \bibinfo{year}{2009}.
\bibitem[{{Cao} et~al.(2019){Cao}, {Hidalgo Martinez}, {Simon}, {Wei}, and
  {Sheikh}}]{cao2019openpose}
\bibinfo{author}{Z.~{Cao}}, \bibinfo{author}{G.~{Hidalgo Martinez}},
  \bibinfo{author}{T.~{Simon}}, \bibinfo{author}{S.~{Wei}},
  \bibinfo{author}{Y.~A. {Sheikh}},
\newblock \bibinfo{title}{{OpenPose}: realtime multi-person {2D} pose
  estimation using {Part Affinity Fields}},
\newblock \bibinfo{journal}{IEEE Transactions on Pattern Analysis and Machine
  Intelligence} \bibinfo{volume}{43} (\bibinfo{year}{2019})
  \bibinfo{pages}{172--186}.
\bibitem[{{Padilla} et~al.(2020){Padilla}, {Netto}, and {da
  Silva}}]{padillaCITE2020}
\bibinfo{author}{R.~{Padilla}}, \bibinfo{author}{S.~L. {Netto}},
  \bibinfo{author}{E.~A.~B. {da Silva}},
\newblock \bibinfo{title}{A survey on performance metrics for object-detection
  algorithms},
\newblock in: \bibinfo{booktitle}{2020 International Conference on Systems,
  Signals and Image Processing (IWSSIP)}, \bibinfo{year}{2020}, pp.
  \bibinfo{pages}{237--242}.

\end{thebibliography}







\end{document}